\documentclass{sbc2023} 
\usepackage{graphicx}
\usepackage[misc,geometry]{ifsym} 
\usepackage{fontspec}
\usepackage{academicons}
\usepackage{color}
\usepackage{hyperref} 
\usepackage{aas_macros}
\usepackage[bottom]{footmisc}
\usepackage{supertabular}
\usepackage{afterpage}
\usepackage{url}
\usepackage{pifont}
\usepackage{multicol}
\usepackage{multirow}
\usepackage{longtable}
\usepackage{xcolor}
\usepackage{subcaption}
\renewcommand{\cite}{\citep}

\setcitestyle{square}

\definecolor{orcidlogo}{rgb}{0.37,0.48,0.13}
\definecolor{unilogo}{rgb}{0.16, 0.26, 0.58}
\definecolor{maillogo}{rgb}{0.58, 0.16, 0.26}
\definecolor{darkblue}{rgb}{0.0,0.0,0.0}
\hypersetup{colorlinks,breaklinks,
            linkcolor=darkblue,urlcolor=darkblue,
            anchorcolor=darkblue,citecolor=darkblue}

\definecolor{darkgreen}{rgb}{0.0,0.5,0.0}

\jid{JBCS}
\jtitle{\Large \textbf{Preprint of Paper accepted for publication in the Journal of the Brazilian Computer Society (JBCS)}}

\jyear{2026}

\title[Graph Neural Networks for Semi-Supervised Image Classification with Multi-Feature Aggregation]{Graph Neural Networks for Semi-Supervised Image Classification with Multi-Feature Aggregation}

\author[Gapski et al. 2026]{

\affil{\textbf{Marina Chagas Bulach Gapski}
~\textcolor{blue}{*}~[~\textbf{S\~{a}o Paulo State University (UNESP), Rio Claro, Brazil~}~|\href{mailto:marina.gapski@unesp.br}{~\textbf{\textit{marina.gapski@unesp.br}}}~]}
\affil{\textbf{Vinicius Atsushi Sato Kawai}~\textcolor{blue}{*}~[~\textbf{S\~{a}o Paulo State University (UNESP), Rio Claro, Brazil}~|\href{mailto:vinicius.kawai@unesp.br}{~\textbf{\textit{vinicius.kawai@unesp.br}}}~]}
\affil{\textbf{Gustavo Rosseto Leticio}~\textcolor{blue}{*}~[~\textbf{S\~{a}o Paulo State University (UNESP), Rio Claro, Brazil}~|\href{mailto:gustavo.leticio@unesp.br}{~\textbf{\textit{gustavo.leticio@unesp.br}}}~]}
\affil{\textbf{Lucas Pascotti Valem}~\textcolor{green}{*}~[~\textbf{University of S\~{a}o Paulo (USP), S\~{a}o Carlos, Brazil}~|\href{mailto:lucas@icmc.usp.br}{~\textbf{\textit{lucas@icmc.usp.br}}~]}}
\affil{\textbf{Daniel Carlos Guimar\~{a}es Pedronette}~\textcolor{blue}{*}~[~\textbf{S\~{a}o Paulo State University (UNESP), Rio Claro, Brazil~}|\href{mailto:daniel.pedronette@unesp.br}{~\textbf{\textit{daniel.pedronette@unesp.br}}}~]}
\affil{\textbf{Mohand Said Allili}~\textcolor{orange}{*}~[~\textbf{Universit\'{e} du Qu\'{e}bec en Outaouais}~|\href{mailto:MohandSaid.allili@uqo.ca}{~\textbf{\textit{mohandsaid.allili@uqo.ca}}}~]}

}

\begin{document}

\begin{frontmatter}
\maketitle

\begin{mail}
Department of Statistics, Applied Mathematics, and Computing (DEMAC), S\~{a}o Paulo State University (UNESP), Av. 24 A, Rio Claro, SP, 13506-900, Brazil.
\end{mail}

\begin{mail2}
Institute of Mathematics and Computer Science (ICMC), University of S\~{a}o Paulo (USP), Av. Trabalhador São-carlense, 400 - Centro, São Carlos, SP, 13566-590, Brazil.
\end{mail2}

\begin{mail3}
Department of Computer Science and Engineering, University of Quebec in Outaouais (UQO), 101 St-Jean Bosco, Gatineau, Quebec,  J8X 3X7,  Canada. 
\end{mail3}

\begin{dates}
\small{\textbf{Accepted:} 08 June 2026}

\end{dates}

\begin{abstract}
\textbf{Abstract.~}
\noindent Feature extraction involves the identification and extraction of salient characteristics or patterns, including edges, textures, shapes, and color attributes. Contemporary feature extractors predominantly leverage deep learning architectures, such as Convolutional Neural Networks (CNNs) and Vision Transformers (VITs). The availability of diverse feature extractors in the literature provides a wide range of feature representations. Features extracted from an image depend on the specific application, the chosen extractor, and its configuration. Therefore, integrating complementary information by combining distinct extractors offers a promising way to enhance performance.
Graph Neural Networks (GNNs), particularly Graph Convolutional Networks (GCNs), have emerged as powerful and widely adopted approaches for semi-supervised image classification, as they effectively leverage both labeled and unlabeled data while exploiting the underlying graph structures that capture relationships among samples.
This study proposes a novel approach for GNNs in scenarios where labeled data is scarce, by integrating diverse sets of feature and graph representations derived from various extractors in classification scenarios. Experimental investigations were conducted, encompassing combinations of distinct feature and graph extractors, as well as rank aggregation strategies. 
The primary contributions of this work are underscored by the experimental findings, which demonstrate that the strategic combination of feature and graph representations, coupled with the application of manifold learning for graph processing, leads to significant improvements in classification accuracy across the majority of experimental conditions. Furthermore, the utilization of rank aggregation techniques to integrate features from different extractors was shown to enhance classification accuracy.

\end{abstract}


\end{frontmatter}

\section{Introduction}
\label{Introducao}

In recent years, the volume of multimedia data has experienced an exponential increase. Daily, millions of individuals share images, videos, and audio across various platforms. Given the vast amount of available content, tasks that were once performed manually have become impractical due to financial, labor, and time constraints~\cite{paperBigDataContrastive}. Consequently, classification methodologies capable of addressing scenarios characterized by limited labeled data, such as semi-supervised and weakly supervised learning, have become increasingly crucial. The acquisition of labeled data is often resource-intensive and time-consuming, necessitating the development of efficient and robust classification strategies. In the domain of image data processing, feature extraction methodologies are widely employed across numerous tasks. 

As the pace of multimedia content creation continues to accelerate, machine learning techniques have emerged as highly promising solutions for classification tasks. Supervised methods can significantly reduce processing time; however, they present a major challenge: the need for large volumes of labeled data. In datasets containing millions of images, this requirement becomes a substantial obstacle. This challenge is central to the effectiveness of semi-supervised and unsupervised learning strategies, which are designed to function with limited or no labeled data.

Semi-Supervised Learning (SSL)~\cite{Semi-Supervised-Learning} occupies a position between supervised and unsupervised learning. In SSL, the algorithm is provided with both unlabeled data and some labeled information, learning from both. In some scenarios, labeled data might be scarce, for example, 20\% or 10\% of the total dataset. Typically, this supervised information consists of labels associated with a subset of the data points~\cite{Semi-Supervised-Learning}. This approach significantly reduces the manual effort required, as only a small proportion of the data needs to be labeled manually. As a result, semi-supervised and unsupervised methods can save considerable time and labor in tasks that were previously reliant on manual processing~\cite{valem2023graph}.

The applications of classification tasks are vast and encompass a wide range of data types, including audio, video, and images. Image datasets containing millions of items, along with frequent updates, stand to benefit significantly from semi-supervised and unsupervised classification techniques. Examples of such datasets include collections of images representing different flower species~\cite{Flowers}, various dog breeds~\cite{dogs}, and car brands~\cite{cars}, among others. These datasets mirror real-world challenges and provide valuable opportunities for evaluating diverse classification strategies. Presently, applications leveraging these strategies are increasingly prevalent in daily life. One example is photo organizing applications that automatically identify people, animals, places, and objects in images, serving millions of users and managing thousands of new entries daily. Social media platforms also stand to gain from the use of neural networks~\cite{ARGYRIS2020106443}.

The wide range of applications has led to the development of numerous models and feature extractors. Convolutional Neural Networks (CNNs)~\cite{paperImagenet, CNN}, which employ various types of layers, such as convolutional, non-linear, and pooling layers, among others, have become a staple in image classification tasks. More recently, Transformer-based models have also been utilized for feature extraction, with notable examples including VIT-B16~\cite{ViT-b16} and SWIN-TF~\cite{Swin-tf}. A promising approach involves combining multiple feature extractors to obtain complementary insights from the same dataset. This is because different extractors evaluate and extract the most relevant features based on their own models, which may result in discrepancies in their interpretations of the data.
 
From an alternative perspective, graph-based models and convolutional networks have been proposed, yielding impressive results. Specifically, a type of Graph Neural Networks (GNNs), the Graph Convolutional Networks (GCNs)~\cite{GCN, gcn-sgc} are capable of performing convolutions in the non-Euclidean domain defined by graphs, demonstrating considerable effectiveness in semi-supervised classification tasks.

Unlike most traditional classifiers, GNNs not only rely on input features but also require a graph structure for learning. Graphs, which represent relationships and connections between entities, are highly versatile and can be used to identify and analyze patterns within data. GCNs are particularly well-suited for semi-supervised classification because they can leverage both labeled and unlabeled data, taking advantage of the graph structure and neighborhood information to enrich the learning process.

Initially, GCNs were designed for datasets where the graph structure is inherently available, such as citation networks~\cite{paperCora2008, GCN}. However, in the context of image datasets, the graph typically needs to be computed beforehand.
Learning effective representations is central to a wide range of machine learning applications~\cite{uelwer2023survey}. Over the past decade, various feature extractors have been proposed, ranging from hand-crafted methods (e.g., color, texture, and shape) to deep learning-based approaches (e.g., CNNs and Vision Transformers). These extractors are employed to compute feature vectors that capture the most relevant and discriminative information in an image.

It is well-established that different extractors can yield complementary information~\cite{paperPirasFusion17}, meaning that combining multiple extractors can lead to enhanced results under certain conditions. However, the challenge lies in determining how to effectively combine these different extractors. The literature suggests two primary strategies for this: early fusion and late fusion~\cite{paperPirasFusion17}. Early fusion involves directly combining features at the initial stages, while late fusion integrates features at higher levels, such as matrices, graphs, or ranked lists.

This work proposes a novel approach that combines different feature extractors. Taking into account the properties of GNN models, which have two distinct inputs (e.g., a set of features and a graph), we leverage this characteristic to integrate graphs and features from various extractors, thus exploiting their complementary strengths. The key contributions of this work to the field of semi-supervised image classification using GNNs are as follows:

\begin{itemize} 
    \item An exploration of the benefits of combining graphs and feature sets from multiple extractors demonstrates that this combination can enhance classification performance by capitalizing on the complementary information each extractor provides. 
    \item The application of manifold learning re-ranking techniques to further improve the input graph for GNNs, particularly when utilizing multiple graphs and feature sets. 
    \item An analysis indicating that GCNs are more sensitive to the input graph than to the feature sets, highlighting the importance of constructing a highly effective graph for optimal performance in these applications. 
    \item An experimental evaluation showing superior classification results in most scenarios, emphasizing the effectiveness of our approach in environments with limited labeled data. 
\end{itemize}

This paper is an extension of previous work~\cite{my-paper} with more experiments, GNN models, and datasets. It also proposes a novel approach for rank aggregation, producing a graph from multiple features. Rank aggregation is a process used to combine multiple rankings of items into a single, consistent order. It aims to find the most representative ranking by considering various input rankings, each potentially provided by different sources or algorithms. The goal is to merge these rankings in a way that minimizes inconsistencies and maximizes agreement across all of them. This work also introduces combinations with rank aggregation techniques between features and graphs, which showed improved results. 

\section{Related work}
\label{sec:related-work}

\subsection{Feature fusion for image classification}

Due to the immense volume of data, semi-supervised and unsupervised methods have been highlighted, especially those utilizing neural networks, due to their ability to create models that require little or no user interaction~\cite{PEDRONETTE-}. The combination of graphs, semi-supervised models, and unsupervised learning methods has proven to be quite promising, especially reciprocal neighbor graphs~\cite{valem2023graph}.

To describe and represent an image in vector form, various types of extractors can be used, for example, transformers~\cite{Swin-tf}~\cite{ViT-b16}, and neural networks (for instance, CNN~\cite{CNN}). These extractors describe the images by extracting numerical characteristics that represent visual features, such as texture, shape, and color, among others. Texture, for example, is a relevant element for human vision and also a crucial characteristic for computer vision models~\cite{featuresTexturas}. 

It is possible that high-dimensional feature extraction may reduce the final accuracy of the model~\cite{rahma2023combination}. To address this issue, there are several methods for reducing feature dimensions. Some of these techniques involve learning the subspace of the extracted features. However, this can lead to overfitting in the subspace, often resulting in precision loss in the model~\cite{JIANG2023108817}. A recent method aims to deal with this issue through a locality-preserving projection, using dual projection matrices, allowing the preservation of important data properties~\cite{JIANG2023108817}. 

Several characteristics can be considered depending on the type of image in the set and also the application~\cite{012028}. The features extracted from images depend on the type of extractor used and its configuration. SWIN-TF is a transformer-based extractor. Traditional transformers are based on global attention calculations. SWIN-TF employs a hierarchical structure commonly used in CNN and performs self-attention calculations in non-overlapping windows. This reduces computational complexity and is highly useful in applications aimed at extracting characteristics from focused and unfocused regions. SWIN-TF can demonstrate good results in such applications~\cite{WANG2023101978}. 

\subsection{GNN-based Semi-Supervised Image Classification}

One of the greatest challenges in image classification is that close images in the feature space do not necessarily have the same semantic content; thus, often, more than the image's content itself needs to be evaluated~\cite{6654170}. Depending on the characteristics extracted by the chosen extractor and the configuration parameters used in the extractor, images with different semantic contents might be considered similar by the classification or ranking model. This difference, though obvious for humans, can pose a significant challenge for the mathematical model. One possible solution is to consider more than just the image content and also take into account their similarity.
When ranking information is represented by structures like graphs and hypergraphs, one can analyze how similar objects will reference each other and have rankings of similar neighbors~\cite{VALEM2022104473}. Thus, unsupervised methods are capable of obtaining better and more accurate rankings, depending on little or no user assistance. For achieving better results in specific scenarios, one can consider specific features from a particular set through ad hoc methods~\cite{10.1093/bib/bbx101}.

The GNN models have shown great potential for semi-supervised learning tasks~\cite{valem2023graph}. Combinations of semi-supervised, unsupervised methods, and neural networks have been promising. Recent studies demonstrate that using unsupervised learning methods as a pre-processing step in re-ranking tasks has shown a great ability to improve GCN accuracy~\cite{valem2023graph}. GCN can also be used with context-sensitive learning methods, employing a multi-scale graph to explore spatial information~\cite{vats2023surveygraphattentionbased}. Thus, the network becomes capable of extracting global and contextual information, enhancing its precision. The graph structure can also be reconstructed using backpropagation techniques as part of the network training to reduce the influence of the initial graph error on the final classification outcome~\cite{9411656}. 

\section{Proposed Approach}
\label{sec:proposed-approach}

\begin{table*}[t!]
\centering
\small
\setlength{\tabcolsep}{6pt}
\renewcommand{\arraystretch}{1.15}
\caption{Summary of the feature extractors used to generate the embeddings for the experiments.}
\resizebox{\textwidth}{!}{%
\begin{tabular}{l l c l l l}
\hline
\textbf{Name} & \textbf{Model Variant} & \textbf{Dimensionality} & \textbf{Extracted Layer} & \textbf{Pretrain} & \textbf{URL Source} \\
\hline
CNN-ResNet    & 152 layers                & 2048 & Penultimate & ImageNet & \href{https://github.com/Cadene/pretrained-models.pytorch}{github.com/Cadene/pretrained-models.pytorch} \\
CNN-SENet     & 154 layers                & 2048 & Penultimate & ImageNet & \href{https://github.com/Cadene/pretrained-models.pytorch}{github.com/Cadene/pretrained-models.pytorch} \\
CNN-DPNet     & 92 layers                 & 1000 & Penultimate & ImageNet & \href{https://github.com/Cadene/pretrained-models.pytorch}{github.com/Cadene/pretrained-models.pytorch} \\
T2T-VIT24 & T2T-ViT-T-24            & 1000 & Last        & ImageNet & \href{https://github.com/yitu-opensource/T2T-ViT}{github.com/yitu-opensource/T2T-ViT} \\
VIT-B16   & ViT-Base (Patch=16)       & 1000 & Last        & ImageNet & \href{https://github.com/faustomorales/vit-keras}{github.com/faustomorales/vit-keras} \\
SWIN-TF   & Swin-Base-224             & 1000 & Penultimate & ImageNet & \href{https://github.com/rishigami/Swin-Transformer-TF}{github.com/rishigami/Swin-Transformer-TF} \\
ConvNeXt  & ConvNeXt-Base-224 & 1024 & Penultimate & ImageNet & \href{https://github.com/facebookresearch/ConvNeXt}{github.com/facebookresearch/ConvNeXt} \\
DINOv2    & DINOv2 VIT-L14 (large)    & 1024 & Last        & LVD-142M & \href{https://github.com/facebookresearch/dinov2}{github.com/facebookresearch/dinov2} \\
\hline
\end{tabular}%
}
\label{tab:backbones_layers}
\end{table*}

This paper extends the work presented in \cite{my-paper}, which addresses the challenge of enhancing image classification techniques in scenarios where labeled data is limited, a common issue in numerous real-world applications. Acquiring labeled data is often resource-intensive and time-consuming, thus methods that can effectively operate with limited labeled data, such as semi-supervised and weakly supervised learning, have gained significant importance. The paper emphasizes the critical role of feature extraction in image classification, where essential image patterns such as edges, textures, and colors are identified. In practice, deep learning models like CNNs and VITs are widely employed for this purpose. 

Given the diversity of feature extractors available, each providing distinct features based on the method and its configuration, combining multiple extractors holds the potential to capture complementary information that can enhance classification performance. Focusing on semi-supervised image classification, the paper investigates the application of GNNs. GNNs are particularly beneficial in scenarios where both labeled and unlabeled data are present, as they leverage graph structures to improve learning outcomes. We propose an innovative approach that integrates multiple feature extraction methods and graph structures to enhance classification accuracy, even when the amount of labeled data is limited. A key contribution of the paper lies in its experimental findings, which demonstrate that the combination of different feature extraction approaches, coupled with the application of manifold learning techniques for graph processing, significantly improves classification performance in most cases.

In this paper, we propose a novel approach that integrates multiple feature extractors via graphs and feature sets, capitalizing on the inherent properties and advantages of GNN models. GNNs are particularly effective in representing information through vector embeddings and graph structures. Our approach explores a variety of techniques for generating representations and constructing graphs, with the aim of combining complementary data from diverse extractors.

In this work, we use the terms descriptor and feature extractor interchangeably~\cite{PaperCBIRTorres}.
Given that different feature extractors capture unique numerical attributes of each image, depending on the specific technique and configuration parameters, our objective is to merge the information from these extractors. In doing so, we aim to assess the impact of this integration on the performance of the GNN, especially regarding classification accuracy. 

Motivated by prior work that manifold learning approaches can improve graph quality for semi-supervised image classification~\cite{valem2023graph}, we investigate whether different combinations of features within a rank aggregation framework influence the classification accuracy of GNN models.

\begin{figure*}[ht!]
\centering
\includegraphics[width=\textwidth]{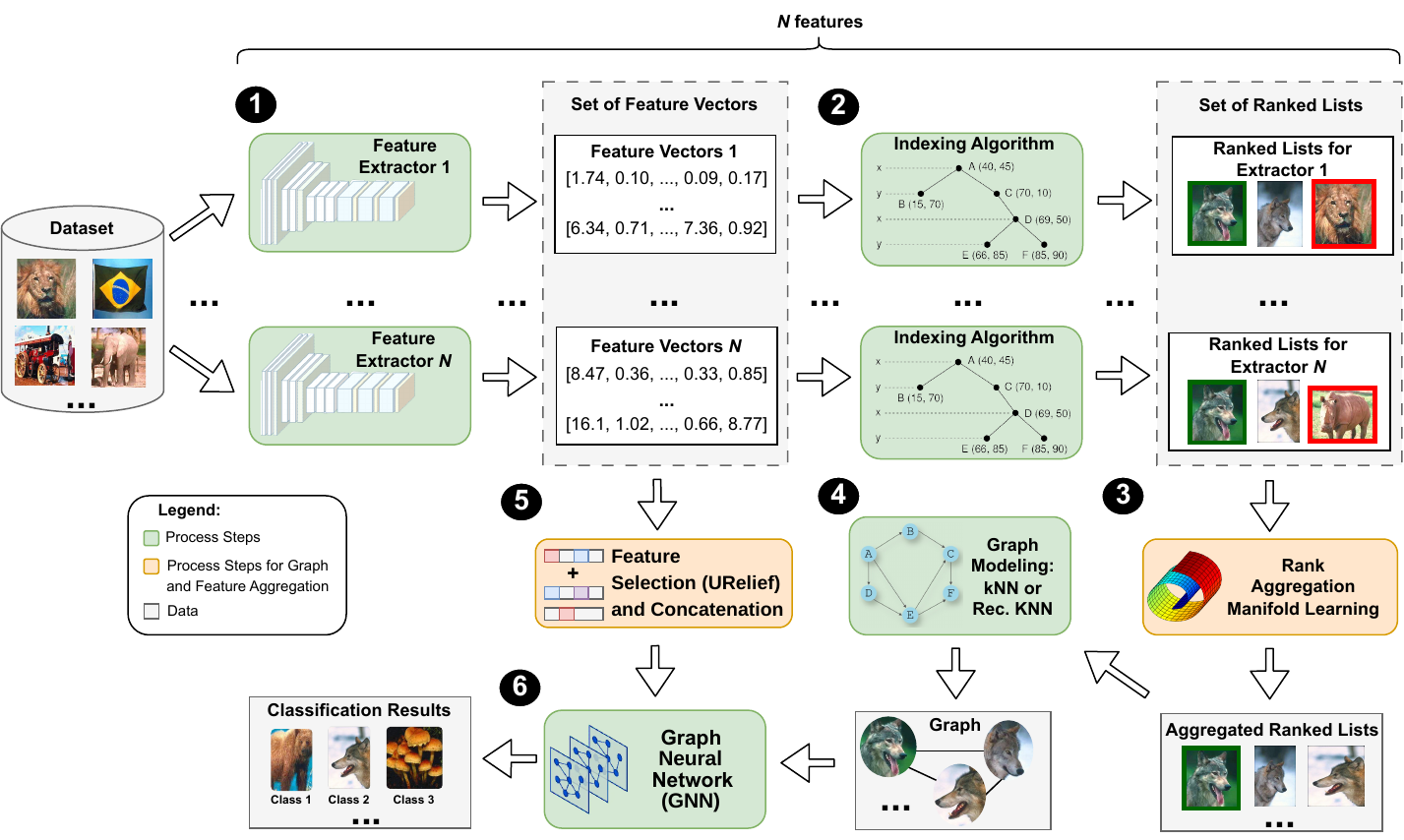}
\caption{Proposed method for combinations between different extractors using manifold learning.}
\label{fig:IlustracaoDois}
\end{figure*}

Although Figure~\ref{fig:IlustracaoDois} depicts a generic setting with $N$ feature extractors, the proposed approach is evaluated under two distinct experimental settings, which define how descriptors are employed for graph construction and node feature generation:

\begin{itemize}
\item \textbf{Setting (i): single-feature or cross-combination.}
In this setting, a single descriptor is used to compute the ranked list and construct the $k$NN graph (Steps~2 to~4), while node features are extracted from one descriptor in Step~1. The descriptor used for graph construction may coincide with or differ from the one used for node features, corresponding to the baseline and cross-combination cases, respectively. When manifold learning is enabled, Step~3 performs only re-ranking, and Step~5 is not applied.

\item \textbf{Setting (ii): multi-feature aggregation.}
In this setting, multiple descriptors are jointly employed. A ranked list is computed independently for each descriptor in Step~2 and then combined using the UDLF rank aggregation module in Step~3, producing a single aggregated ranking. This ranking is used to construct a reciprocal $k$NN graph in Step~4. In parallel, node features are obtained by applying Unsupervised Relief (URelief) to each descriptor and concatenating the selected components in Step~5. The resulting graph and fused node features are then used to train the GNN in Step~6.
\end{itemize}

Notice that Steps~3 and~5, highlighted in orange, correspond to the main contributions introduced regarding our previous work~\cite{my-paper}. These steps enable the integration of multiple descriptors by, respectively, performing rank aggregation (Step~3) and fusing multiple feature representations into a single node feature vector (Step~5).

Overall, this workflow enables a comparison between using a single extractor for both node feature representation and graph construction (i.e., edge definition), employing distinct extractors for each component, and jointly aggregating multiple descriptors. Moreover, it allows us to evaluate the impact of different manifold learning strategies and feature combinations within the rank aggregation process on the resulting classification performance.

\subsection{Feature Extraction}

The workflow begins with an image dataset used across all processing steps. From it, features are extracted using $N$ feature extractors. For each image in the dataset, the extractor processes the image and generates a feature vector encoding its relevant characteristics and patterns. 

A feature matrix is created for the set. Each row of the matrix is the feature vector of an image in the dataset. Each element of this matrix, $f_{i, j}$, represents a numerical value, where row $i$ represents an image and $j$ represents a specific feature.

To facilitate reproducibility of the experimental setup, Table~\ref{tab:backbones_layers} provides information about the feature extractors used in our experiments, including the model variants (e.g., base or large, and the number of layers), feature dimensionality, the specific layer from which features are extracted, and the source URLs for both the implementation and the pretrained weights. This information allows the experiments to be faithfully reproduced and fairly compared.
All backbones were used as frozen, pretrained feature extractors. No fine-tuning or additional training was performed by us on any of them. The features were extracted directly from the pretrained models.
A detailed description of the architecture of each backbone is provided below:

\begin{itemize}
    \item \textbf{CNN-ResNet~\cite{he2015deepresiduallearningimage}}: CNN-ResNet is a convolutional neural network that uses residual blocks. The activation functions for some layers are connected, and the model has shortcuts that allow some layers to be skipped. CNN-ResNet learns by combining these residual blocks. So, the network is adjusted only in the residual mapping~\cite{he2015deepresiduallearningimage}. This work considers the implementation with 152 layers.
    \item \textbf{CNN-DPNET~\cite{dpnet}}: The Dual Path Network (DPN) is a highly efficient and modularized image classification architecture that combines the advantages of Residual Networks (ResNet) and Densely Convolutional Networks (DenseNet) by using a new internal connection topology. DPN leverages ResNet’s ability to reuse features and DenseNet’s capacity to explore new features, allowing for the learning of more robust representations. The network maintains a flexible dual-path structure that shares common features while also enabling the exploration of new ones.  
    \item \textbf{CNN-SENet~\cite{senet}}: CNN-SENet is a type of CNN that incorporates a Squeeze-and-Excitation (SE) block to improve feature representation. The SE block recalibrates channel-wise feature responses by modeling the interdependencies between channels, allowing the network to focus on the most important features. This approach boosts performance significantly with minimal computational cost and is able to achieve excellent results. 
    \item \textbf{T2T-VIT~\cite{t2t}}: T2T-ViT (Tokens-to-Token Vision Transformer) is a variant of the VIT designed to address the limitations of VIT when trained from scratch on datasets like ImageNet. It improves upon VIT by introducing a Tokens-to-Token transformation, which progressively aggregates neighboring tokens to better capture local structures (such as edges and lines) within the image. This transformation helps improve training efficiency and reduces token length.  
    \item \textbf{Vision Transformer (VIT)~\cite{ViT-b16}}: A Transformer pre-trained on large amounts of data and transferred to various image recognition benchmarks. First, the image is divided into patches, which are flattened and mapped so that they all have the same dimensions. Then, lower-dimensional linear embeddings are produced from the flattened patches. Positional embeddings are added, and the input sequence is fed as it would be for a standard transformer encoder. Finally, the model is pre-trained with image labels in a supervised manner using a very large dataset. Afterward, transfer learning is performed on the target dataset for feature extraction. The base version with a patch size equal to 16 was used in this work.
    \item \textbf{Shifted windows Transformer (SWIN-TF)~\cite{Swin-tf}}: It uses a Vision Transformer called Swin Transformer, which serves as a general-purpose backbone for computer vision. The Transformer is hierarchical, and its representation is calculated with \textbf{S}hifted \textbf{win}dows~\cite{Swin-tf}. The shifted windows scheme allows to achieve great efficiency by limiting self-attention computation to non-overlapping local windows while allowing connections between windows. This hierarchical architecture is flexible for modeling at various scales and has linear computational complexity with respect to the image size.
    \item \textbf{DINOv2~\cite{dinov2}}: DINOv2 is a series of image encoders pretrained on large curated datasets without supervision. DINOv2 models exhibit an understanding of object parts and scene geometry across various image domains. These visual features can be easily used with simple classifiers like linear layers, and the work envisions further scaling to create AI systems capable of processing visual features like word tokens in language models. 
    \item \textbf{ConvNeXt~\cite{convnext}}: ConvNeXt is a family of modernized Convolutional Neural Networks (ConvNets) designed to compete with Transformers in visual recognition tasks. By reexamining and gradually adapting a standard CNN-ResNet toward a Transformer-like architecture, ConvNeXt incorporates several key design improvements that boost performance.  
\end{itemize}

The same set of descriptors is used across all datasets. The only exception is DINOv2, which was excluded from the Flowers and CUB200 datasets due to potential overlap with its pretraining data. Therefore, this descriptor is used only for the Corel5k dataset.

\subsection{Indexing Algorithm} 

The similarities between images are computed based on the features obtained in Step 1 of Figure \ref{fig:IlustracaoDois}. For each feature extractor, a corresponding ranked list is generated.
This can be efficiently performed by optimized indexing algorithms, such as the BallTree~\cite{PaperBallTree}.

For this work, this process is used to obtain a ranked list, which is a list that contains the nearest elements for each element in the set, ranked from closest to the most distant one (i.e., in descending order of similarity).
This list is generated using the Euclidean distance, which calculates the distance between each image and every other image in the set. 

\subsection{Re-Ranking and Rank Aggregation by Manifold Learning} 

The ranked lists (obtained in step 2) are processed by an unsupervised similarity learning method, which aims at post-processing and improving the effectiveness of the ranked lists. The same methods can be used to do rank aggregation (step 3 in Figure \ref{fig:IlustracaoDois}). 

Manifold learning approaches aim to capture and exploit the intrinsic manifold structure to compute a more effective distance/similarity measure~\cite{paperManifoldDef2011}. There are different methods in the literature that perform unsupervised similarity learning considering different strategies (e.g., graphs, diffusion processes, and others).

In this work, we consider unsupervised manifold learning methods to provide more effective similarity measures using rank-based formulations. They are all implemented on the Unsupervised Distance Learning Framework (UDLF)~\cite{UDLF}, which is an open-source framework that provides different approaches in this category. The manifold learning methods used in this work are listed below: 

\begin{itemize}
    \item \textbf{Breadth-First Search Tree (BFSTree)~\cite{BFSTree}}: Utilizes a breadth-first tree to represent similarity information provided by ranking references. This tree is exploited to discover underlying similarity relationships, enabling a more effective similarity measure to be computed.

    \item \textbf{Rank-Based Diffusion Process with Assured Convergence (RDPAC)~\cite{rdpac}}: An efficient rank-based diffusion process combining diffusion approaches and ranking-based strategies.
    It approximates a diffusion process by leveraging rank-based information while ensuring convergence. It is a low asymptotic complexity algorithm and can be computed regionally.

    \item \textbf{Log-based Hypergraph of Ranking References (LHRR)~\cite{lhrr}}: It is a hypergraph representation that, along with its corresponding hyperedges, is grounded in Ranking References, which are assigned weights based on a logarithmic function. Hypergraphs serve as a potent extension of traditional graphs, enabling the definition of hyperedges that can link any number of vertices. This capability allows for the representation of similarity relationships among sets of objects, as opposed to merely pairwise connections typically found in conventional graph-based models. Identifying sets of similar objects is of paramount importance for the task at hand.
    
\end{itemize}

\subsection{Graph modeling}

This step is essential, as in most image datasets the graph is not readily available, and a strategy to construct it is crucial, potentially affecting the final results. Using the ranked lists obtained in Step 2, which are refined by manifold learning techniques, we build a k-nearest neighbors (kNN) graph to serve as input for the GNN. In this graph, each image is represented as a node, and edges are established from each node to its top-$k$ most similar elements according to the corresponding ranked list.

Alternatively, a reciprocal kNN graph can be constructed by adding a constraint: an edge between two nodes exists only if both appear among each other's $k$ nearest neighbors~\cite{valem2023graph}. This results in a symmetric graph, where all connections are bidirectional, reinforcing mutual similarity.

In this work, we adopt a reciprocal kNN graph when multiple features are used.
This is part of the novelty of our approach: by applying the reciprocity constraint in the multi-feature setting, we enhance the integration of complementary information while maintaining strong and meaningful connections in the graph structure.

\subsection{Feature Selection and Concatenation}

Feature matrices typically have high dimensionality. Each extracted numerical characteristic from an image results in a new dimension, which can lead to very large matrix representations. 

In the present work, URelief is applied to each feature set independently to select 200 features. These features are then concatenated to obtain the final representation used as input to the GNN, which results in a lower-dimensional representation, in comparison to the original representation.

After applying URelief to each feature vector, the low-dimensional vectors from multiple descriptors are then concatenated into a single final vector. This concatenation combines complementary information from different descriptors and is used as input for the GNN model, as illustrated in Figure~\ref{fig:IlustracaoDois} steps 5 and 6.

\subsection{Graph Neural Network (GNN)}

A Graph Convolutional Network (GCN)~\cite{GCN} is a type of Graph Neural Network (GNN) that performs convolutional operations on graph-structured data for classification tasks. A key advantage of GCNs is their ability to exploit the graph connectivity to propagate and smooth information across neighboring nodes, computing new representations (i.e., embeddings) while remaining computationally efficient. In this work, we use the implementation provided by the PyTorch Geometric framework~\cite{fey2019fastgraphrepresentationlearning,NEURIPS2019_bdbca288}\footnote{PyTorch Geometric: \url{https://github.com/pyg-team/pytorch_geometric}.}.
We employ graph neural network (GNN) models as semi-supervised, transductive image classifiers that leverage relational information encoded in graph structures. The different GNN models considered in this work are described below:

\begin{itemize}
    \item \textbf{Graph Convolution Network (GCN)}~\cite{GCN}: The pioneering GCN that introduced the concept of applying convolutions to graph structures, commonly referred to as GCN.
    \item \textbf{Approximate Personalized Propagation of Neural Predictions (APPNP)}~\cite{appnp}: A model that integrates a GCN with the PageRank algorithm, utilizing a propagation strategy derived from a modified PageRank approach. 
    \item \textbf{Graph Attention Networks (GATs)~\cite{gcn-gat}}: a neural network architecture designed for graph-structured data. GATs leverage masked self-attention layers to overcome limitations inherent in prior graph convolutional methods and their approximations. By stacking attention-based layers, nodes in the graph can selectively attend to different features in their neighborhoods, allowing for the assignment of varying weights to neighbors without the need for computationally expensive matrix operations, such as inversion. 
    \item \textbf{Simple Graph Convolution (SGC)~\cite{gcn-sgc}}: SGC is a GCN with reduced complexity, through successively removing nonlinearities and collapsing weight matrices between consecutive layers. The simplifications done did not impact negatively the accuracy of many downstream applications. 
    \item \textbf{Auto-Regressive Moving Average (ARMA)~\cite{gcn-arma}}: a graph convolutional layer inspired by auto-regressive moving average (ARMA) filter. It is able to capture global graph structures and be robust to noise. ARMA is a graph neural network implementation of the ARMA filter with a distributed and recursive formulation. It results in a convolutional layer that is localized in the node space, is efficient to train, and can be transferred to new graphs at test time. 
\end{itemize}

In this work, we train the GNN using (i) node features and (ii) a graph built from ranked lists refined by manifold learning.
In the \textbf{single-feature/cross-combination} setting, the descriptor used to construct the graph may be the same as (baseline) or different from (cross-combination) the descriptor used to compute node features.
In the \textbf{multi-feature} setting, multiple descriptors are jointly used: ranked lists are aggregated to build a single reciprocal $k$NN graph, while node features are fused via URelief+concatenation, as shown in Steps~3--5 of Figure~\ref{fig:IlustracaoDois}.

\subsection{Overview of the Execution Flow}
\label{subsec:overview_steps}

Given the individual explanation of each step, we now provide an overview of how these steps are connected within the workflow and how they are used in our experiments.

Our approach is divided into two different experimental settings: \emph{(i)} experiments in which a single backbone feature extractor (descriptor) is used to generate both the node features and the graph structure. We refer to this setting as single-feature or cross-combination; and
\emph{(ii)} experiments in which multiple backbone feature extractors (descriptors) are jointly used to construct the graph and the node features.

For setting~(i), the executed steps are as follows, following the numbering in Figure~\ref{fig:IlustracaoDois}:
\begin{enumerate}
    \item Feature extraction is performed, with at most two descriptors, one used to construct the graph and the other to define the node features.
    \item A ranked list is always computed using Euclidean distance and a BallTree structure to obtain the neighborhood information required to build the $k$NN graph.
    \item Manifold learning may or may not be applied. The evaluation explicitly distinguishes these cases, and when used, one of the following methods is adopted: LHRR, BFSTREE, or RDPAC.
    \item The graph is modeled using a standard $k$NN strategy with $k=40$.
    \item This step is not performed.
    \item The resulting graph and feature matrix are then provided to a GNN model, which is trained to perform image classification.
\end{enumerate}

For setting~(ii) the main difference is that both the graph and the image features can be obtained from multiple descriptors, which characterizes the multi-feature setting.
For setting~(ii), the executed steps are as follows, following the numbering in Figure~\ref{fig:IlustracaoDois}:
\begin{enumerate}
    \item Multiple feature extractors are applied, according to the evaluation configuration.
    \item A ranked list is computed independently for each feature using Euclidean distance and a BallTree structure.
    \item Since graph construction requires a single ranked list, all ranked lists are jointly processed using the rank aggregation module of the UDLF framework. This module receives two or more ranked lists, applies manifold learning, and returns a single aggregated ranked list.
    \item A reciprocal $k$NN graph is then constructed, in which only mutual edges are preserved, based on the aggregated ranked list produced by the manifold learning method.
    \item This step is exclusive to the multi-feature setting and is responsible for combining features from multiple extractors using URelief, resulting in a single feature vector per image.
    \item Finally, the GNN model is executed using the resulting graph and feature representation.
\end{enumerate}

\section{Experimental Evaluation} 
\label{sec:experimental-evaluation}

The experimental protocol is divided into four subsections.
While Section~\ref{subsec:experimental-protocol} describes the datasets, parameter settings, and other experimental details; Section~\ref{subsec:ablation_study} presents the ablation study.

We conducted two main sets of experiments. The first set, which is presented in Section~\ref{subsec:single_feature}, investigates the single-feature setting, in which a single backbone descriptor is used to define the node features and, optionally, a different descriptor is used to construct the graph structure. This setting allows us to evaluate both standard single-descriptor configurations and cross-combinations between feature and graph extractors, as well as the impact of applying manifold learning.

The second set, which is presented in Section~\ref{subsec:multi_feature}, investigates the multi-feature setting, in which multiple backbone descriptors are jointly exploited to build both the graph structure and the node features. In this case, ranked lists obtained from different descriptors are aggregated through manifold learning to generate a unified graph, while the corresponding feature representations are fused into a single vector per image. This setting evaluates the benefits of combining complementary visual information from multiple extractors within the proposed framework.

\subsection{Experimental Protocol}
\label{subsec:experimental-protocol}

In this work, we evaluate the proposed approach on three widely used and publicly available image classification datasets:

\begin{itemize}
    \item \textbf{Flowers (Oxford17Flowers)}~\cite{Flowers}: 
    A dataset consisting of 1,360 images grouped into 17 flower categories. 
    Despite its relatively small size, it is commonly adopted in the evaluation of visual classification methods due to its well-defined classes and controlled variability. The average size of the images is around 583 $\times$ 555.

    \item \textbf{Corel5k}~\cite{corel5k}: 
    A multi-class image dataset containing 5,000 images organized into 50 categories. 
    The dataset presents significant intra-class variability and has been extensively used in image retrieval and classification tasks, serving as a standard benchmark in the literature. The average size of the images is around 170 $\times$ 144.

    \item \textbf{CUB200 (CUB-200-2011)}~\cite{cub200}: 
    A fine-grained image classification dataset composed of 11,788 images from 200 bird species. 
    This dataset is particularly challenging due to the high inter-class similarity and subtle visual differences between classes. The average size of the images is around 468 $\times$ 386.
\end{itemize}

For feature extraction, all images were resized to $224 \times 224$, matching the input resolution used to train the feature extractors on ImageNet~\cite{deng2009imagenet} or LVD-142M~\cite{dinov2}.

For manifold learning approaches, the default parameters of the UDLF framework (version 1.60) were used. Since the goal of the present paper is to evaluate the combinations between different feature extractors and graphs, the default parameters were kept at the default values since they are out of the scope of this study. Also, keeping the default values allows us to analyze the results of the combinations with the same parameters without the influence of other aspects on the results.

The only modified parameters were the neighborhood sizes and the maximum list depth \( L \). To ensure a more precise description of the experimental design, we distinguish between \( K_g\), which defines the neighborhood size for graph construction, and \( K_m\), which determines the neighborhood range used by manifold learning methods. In all experiments reported in this work, we set \( K_g = K_m \) to facilitate comparison with baselines and maintain consistency with previous protocols in the literature \cite{valem2023graph, leticio2025neighbor}.

Specifically, we used \( K_g = K_m = 40\) for single-feature experiments and \( K_g = K_m = 50\) for multi-feature aggregation scenarios. The latter choice is justified by the need to exploit the increased complementary information provided by multiple input rankings. Regarding the parameter \( L \), it was set to the dataset size for the Flowers dataset and to 2000 for the remaining datasets. Although \( L \) defines the maximum depth of the ranked lists considered, it does not directly influence the final results as long as \( L > K_g\) and \( L > K_m\), as the similarity relationships and manifold structures are effectively computed within the top-\(K\) range, where the effectiveness typically reaches a plateau \cite{lhrr, BFSTree, rdpac}.

Regarding the hyperparameters used to train the GNN models, the following settings are adopted across all experiments:
\begin{itemize}
    \item Graph neighborhood: $k=40$ for single-feature graph and $k=50$ for multi-feature graph.
    \item Learning Rate: 0.001 for all datasets; except 0.01 for CUB200.
    \item Number of neurons in hidden layers is 64 for all GNNs, except for SGC, which has no hidden layers.
    \item 200 epochs.
    \item Adam optimizer with weight decay of $5 \times 10^{-4}$.
\end{itemize}

All experiments employed a 10-fold cross-validation protocol, where each execution considered one fold as the training set and the others as the testing set, such that each fold is used as the training set at least once. This setting is important because it uses only 10\% of the data for training and 90\% for testing, creating a challenging scenario with limited labeled data and few labeled samples per class. 
Each reported mean and standard deviation were computed over repeated executions of the full cross-validation protocol. Statistical analyses were performed using paired accuracy values obtained under the same folds and random seeds. In some cases, we also report the Wilcoxon signed-rank test $(p < 0.05)$ and Cohen's $d$ to assess the statistical significance and effect size of the observed accuracy differences.

For tables containing a large number of reported values, the best-performing result in each table is consistently highlighted in blue in order to facilitate visual identification and improve the readability and interpretation of the comparative results.

\subsection{Ablation Study}
\label{subsec:ablation_study}

\begin{figure*}[!ht]
\centering

\includegraphics[width=.6\textwidth]{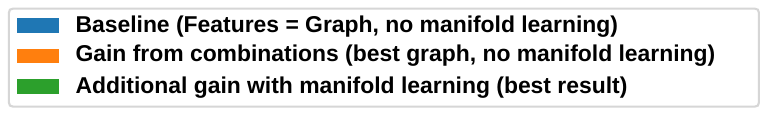}

\vspace{0.4em}

\begin{subfigure}[t]{0.88\textwidth}
\centering
\includegraphics[width=.8\linewidth]{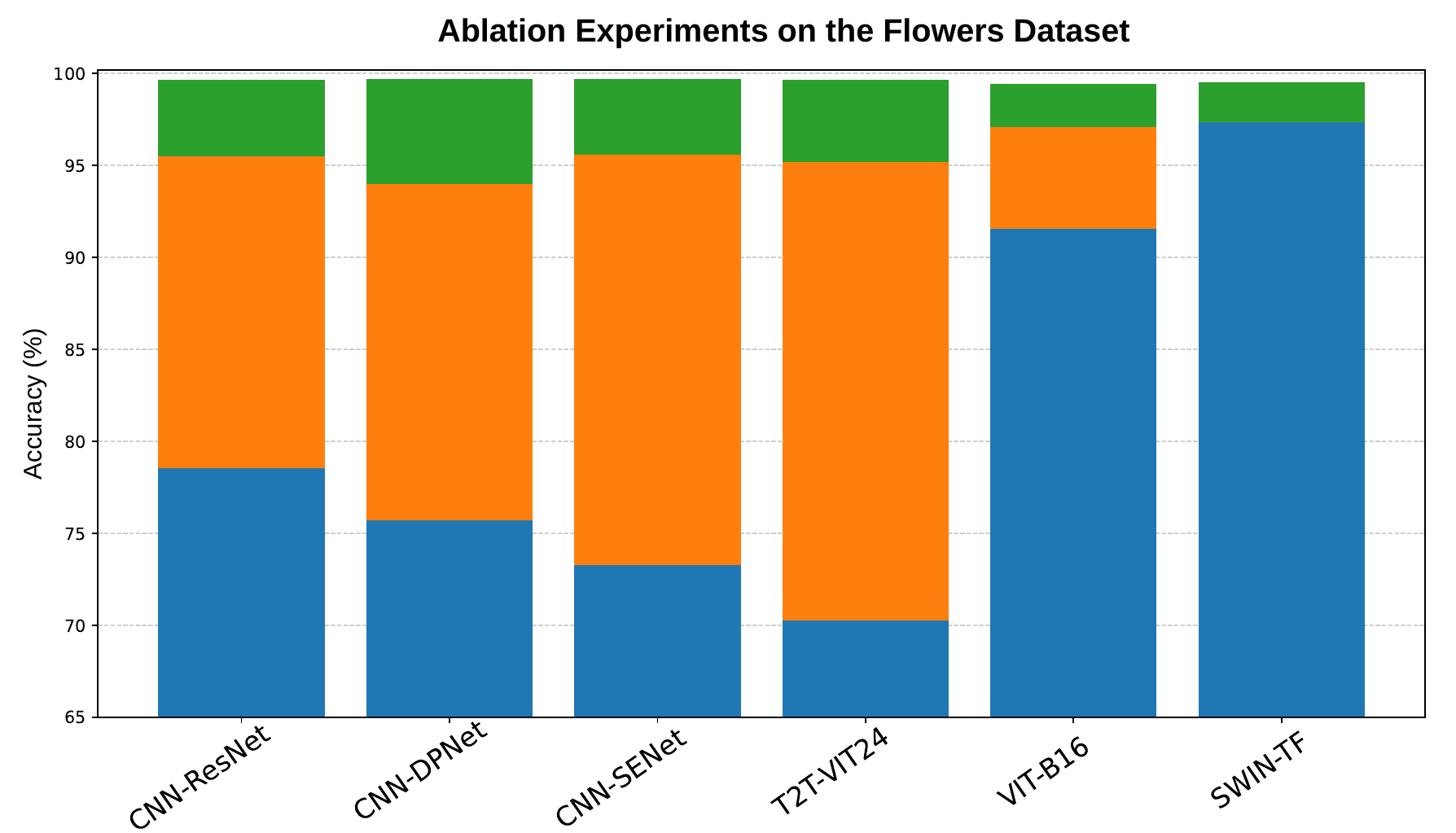}
\caption{Flowers}
\label{fig:ablation_flowers}
\end{subfigure}

\vspace{0.4em}

\begin{subfigure}[t]{0.88\textwidth}
\centering
\includegraphics[width=.8\linewidth]{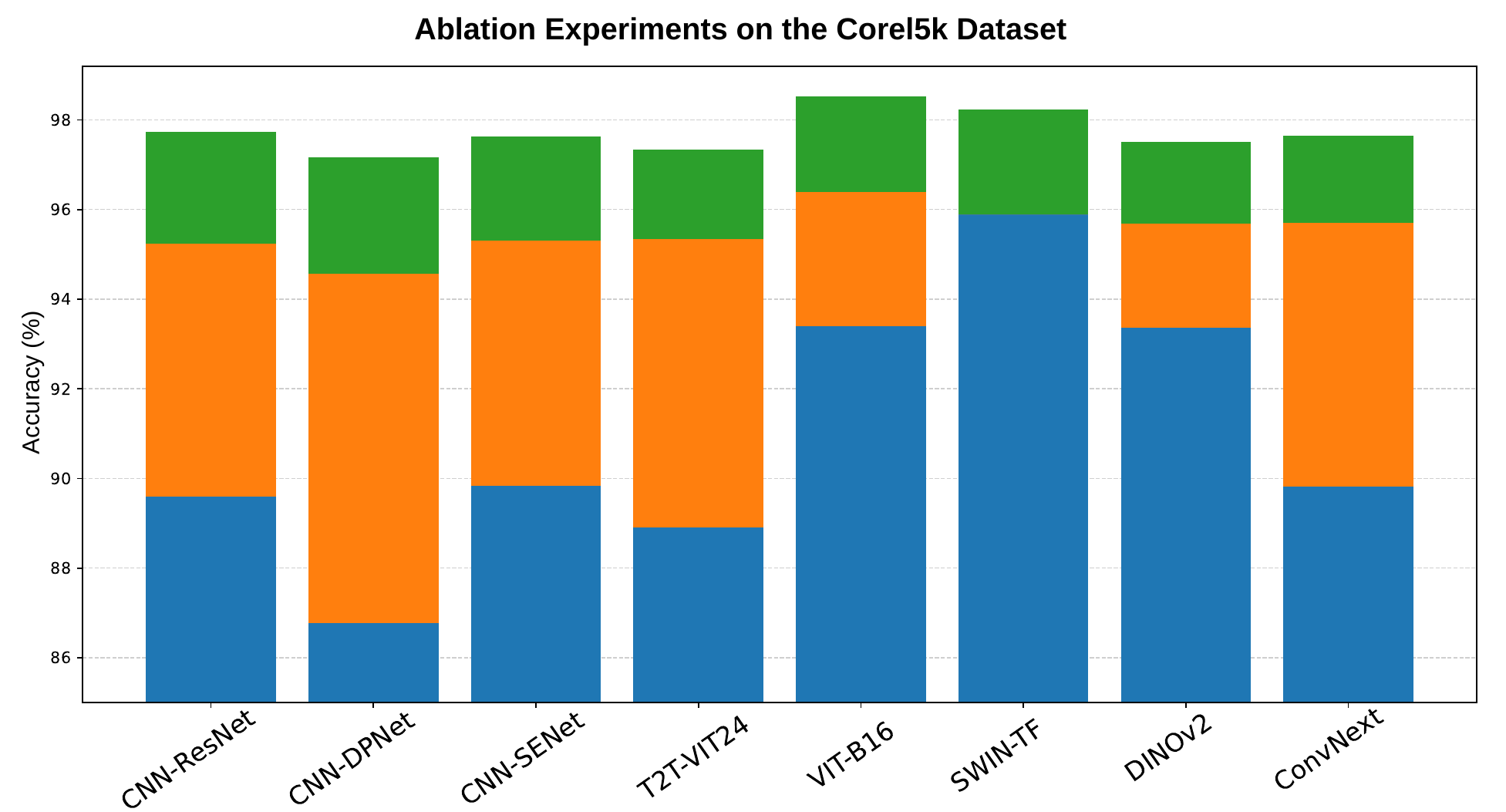}
\caption{Corel5k}
\label{fig:ablation_corel5k}
\end{subfigure}

\vspace{0.4em}

\begin{subfigure}[t]{0.88\textwidth}
\centering
\includegraphics[width=.8\linewidth]{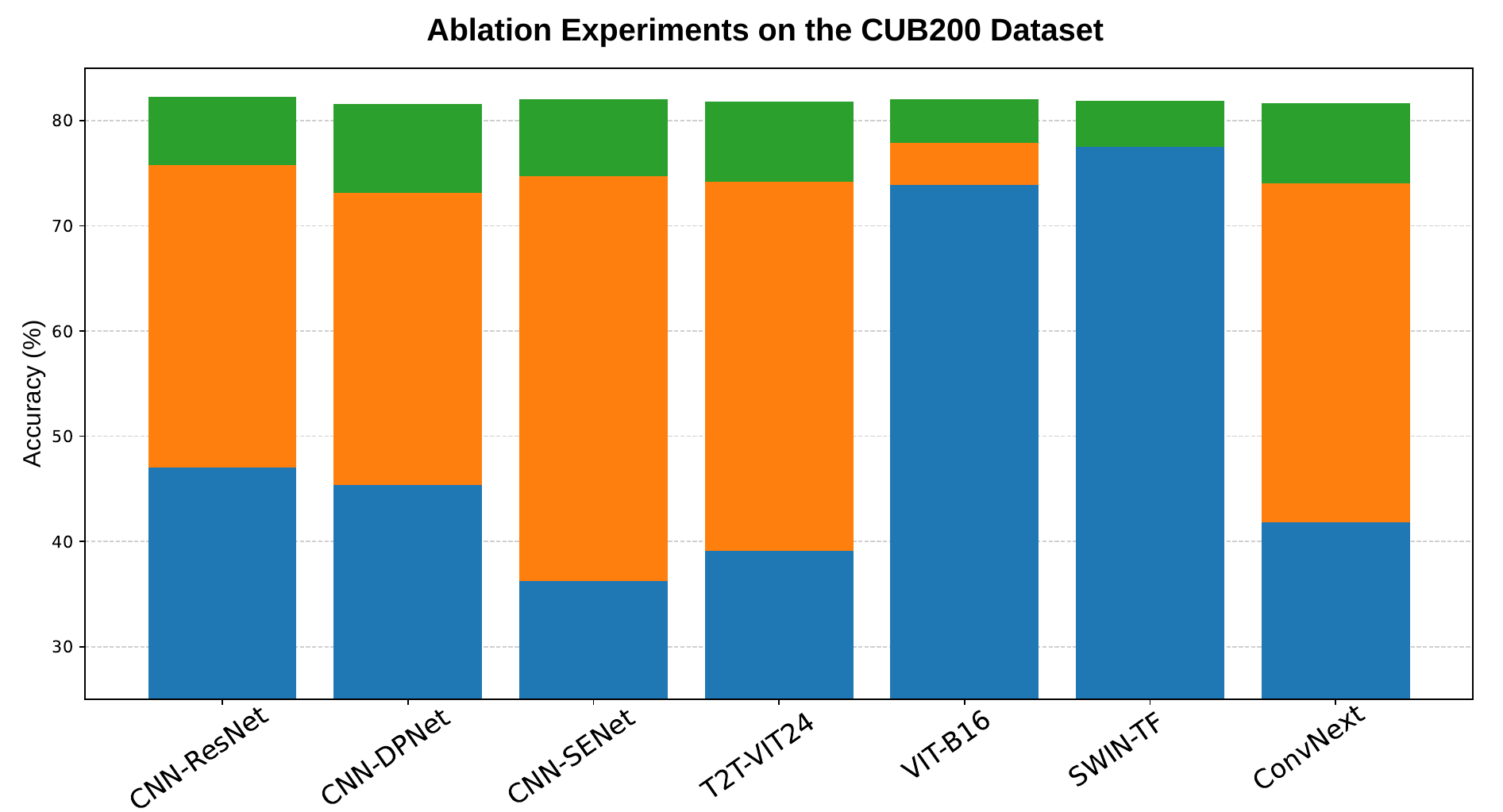}
\caption{CUB200}
\label{fig:ablation_cub200}
\end{subfigure}

\caption{Ablation study on three datasets and multiple features on the SGC model.}
\label{fig:ablation}
\end{figure*}

Since the proposed approach is composed of several components, we conducted an ablation study on each dataset. The results are presented as stacked bar charts for each feature extractor, as shown in Figure~\ref{fig:ablation}. 

Please note that three colors are used in the stacked bars. Blue represents the baseline accuracy when the same feature extractor is used both for the node features and for graph construction. Orange indicates the accuracy gain obtained by changing the graph construction, corresponding to cross-combinations between different feature extractors. Finally, green represents the additional improvement achieved when both graph combination and manifold learning are applied. Since three manifold learning methods were evaluated, the green segment corresponds to the best result among them.

Notice that, typically, the largest gains are obtained when using cross-combinations of features and graphs. In contrast, for the best isolated feature (SWIN-TF), the improvements are more noticeable when feature and graph combinations are jointly employed with manifold learning.

\subsection{Single-Feature Experiments}
\label{subsec:single_feature}

This subsection analyzes how heterogeneous combinations of feature and graph descriptors affect the performance of SGC across different image classification datasets.
Table~\ref{tab:results-sgc-flowers} reports the classification accuracy obtained on the Flowers dataset using the SGC model under different combinations of feature extractors for node features and graph construction, as well as different manifold learning strategies. Overall, the results clearly indicate that combining heterogeneous feature extractors consistently improves performance when compared to single extractor settings, especially when coupled with manifold learning.

First, considering the baseline scenario without manifold learning, it can be observed that the configurations where the same extractor is used for both features and graph construction do not yield the best results in most cases. In contrast, cross-combinations, particularly those involving stronger transformers such as SWIN-TF and VIT-B16, already lead to noticeable accuracy gains.

Similarly to Flowers, Table~\ref{tab:results-sgc-corel5k} confirms that configurations using the same descriptor for both node features and graph construction are rarely optimal on Corel5k. In contrast, combinations that use different descriptors for features and graph consistently achieve higher accuracy, and manifold learning further improves accuracy in most cases. In particular, using SWIN-TF to construct the graph yields strong and stable results across different feature extractors. The best Corel5k performance is obtained with VIT-B16 node features and a SWIN-TF graph under RDPAC, reaching \(98.53 \pm 0.0090\), highlighting the benefit of a transformer-based graph backbone. Even with the inclusion of DINOv2, the strongest results typically rely on heterogeneous feature and graph descriptors, suggesting complementary information between representations.

On the more challenging CUB200 dataset (Table~\ref{tab:results-sgc-cub200}), absolute accuracies decrease as expected due to the fine-grained nature of the task, but the same pattern persists. Settings that use different descriptors for features and graph, combined with manifold learning, provide gains over configurations that rely on a single descriptor for both. Once again, graphs built from SWIN-TF dominate the best results, with the highest accuracy achieved under RDPAC (\(82.25 \pm 0.0227\)), indicating that high-quality similarity neighborhoods become increasingly important as task difficulty increases.

Overall, the joint analysis of Tables~\ref{tab:results-sgc-flowers}, \ref{tab:results-sgc-corel5k}, and \ref{tab:results-sgc-cub200} demonstrates that using different descriptors for node features and graph construction consistently improves SGC accuracy and that manifold learning, particularly RDPAC, further amplifies these gains across datasets.

Given the large number of results reported in these tables, Table~\ref{tab:all_datasets_acc_gain_sgc} was introduced to provide a concise summary. This table compares the best single descriptor results with the best cross-feature combinations for each manifold learning method and dataset, explicitly highlighting the relative gains achieved by combining features and graphs derived from different extractors. This view facilitates the interpretation of the experimental results and enables a clearer comparison across datasets and configurations.

The analysis of single feature cross-combinations directly supports the hypothesis of this work. Although modern transformer based descriptors such as SWIN-TF achieve strong accuracy when used in isolation, its role as graph backbones becomes even more relevant when combined with heterogeneous node features. As shown in Table~\ref{tab:all_datasets_acc_gain_sgc}, cross feature configurations consistently outperform the best single descriptor baselines across all datasets and manifold learning strategies, revealing the complementary nature of different representations. In all cases, the strongest standalone descriptor naturally emerges as the most effective choice for graph construction, reflecting the sensitivity of SGC to graph topology quality. Moreover, manifold learning, particularly RDPAC, further amplifies these gains.

In addition, a statistical analysis was conducted for each row of the table to assess the significance of the observed improvements using two complementary tests: the Wilcoxon signed-rank test $(p < 0.05)$ and Cohen’s $d$. The Wilcoxon signed-rank test was used to evaluate statistical significance, where lower $p$ values indicate stronger evidence against the null hypothesis. Cohen’s $d$ was employed to quantify the magnitude of the improvements. According to standard guidelines, values around $0.2$ correspond to small effects, around $0.5$ to medium effects, and values above $0.8$ to large effects. Overall, the results demonstrate not only statistically significant improvements but also effect sizes that are consistently meaningful.

\subsubsection{Evaluation of Different GNN Models}

Besides SGC, we also evaluate the proposed approach across multiple GNN architectures, namely GCN, GAT, APPNP, and ARMA, under the same experimental setting. Tables~\ref{tab:flowers-networks}, \ref{tab:corel5k-networks}, and \ref{tab:cub200-networks} report the results for the Flowers, Corel5k, and CUB200 datasets, respectively. In these experiments, the graph structure is fixed using SWIN-TF, which achieves the best performance when used in isolation. Importantly, even though SWIN-TF is already the strongest standalone backbone, its performance is further improved when combined with heterogeneous node features and manifold learning strategies. By varying node features and manifold learning while keeping the graph fixed, this setup allows us to isolate the effects of feature heterogeneity and manifold learning from architectural differences.

Across all datasets, the models exhibit a consistent trend. Cross-feature configurations outperform single-feature baselines, and the inclusion of manifold learning further improves performance, with RDPAC achieving the best results in all of these cases. Although absolute accuracy values vary across architectures, the relative performance trends remain stable.
To summarize the best results from Tables~\ref{tab:flowers-networks}, \ref{tab:corel5k-networks}, and \ref{tab:cub200-networks}, we also include Table~\ref{tab:best_gnn_models_with_gains}. As a baseline, we consider results obtained using the same backbone for both feature extraction and graph construction, without manifold learning.
The highest accuracies are achieved by cross-combinations that use different backbones for feature extraction and graph construction, combined with manifold learning (abbreviated as M.L.). Notably, performance gains are observed in all cases. Despite variations in accuracy across models, SGC consistently achieves the best results across all datasets. For SGC, the highest accuracy in each table always corresponds to a cross-feature combination.

These findings reflect the main objective of our approach: demonstrating that combining graphs and features derived from different backbones through manifold learning leads to improved accuracy compared to single-feature setups, which remain the standard in most existing literature, to the best of our knowledge. Consequently, we use SGC for the remaining experiments, as it is both lightweight and competitive. A possible explanation for SGC’s superior performance is its simplified propagation mechanism, which mitigates over-smoothing and enables more effective exploitation of the combined features and the manifold learning stage.

\subsubsection{Comparison with ManifoldGCN}

A comparison between our approach and ManifoldGCN~\cite{valem2023graph} is presented in Table~\ref{tab:summary_manifold_vs_combination_all}. ManifoldGCN is a method that combines GNNs with manifold learning but does not perform feature combinations and has previously been evaluated against state-of-the-art techniques. Both methods use the same settings and graph configuration (i.e., a standard kNN graph). While ManifoldGCN reports results for each descriptor individually and does not support feature combinations, our approach demonstrates that combining the evaluated descriptors can effectively improve performance. The results obtained by our method correspond to the best-performing configurations (highlighted in blue) in Tables~\ref{tab:results-sgc-flowers}, \ref{tab:results-sgc-corel5k}, and \ref{tab:results-sgc-cub200}.

\subsection{Multi-Feature Experiments}
\label{subsec:multi_feature}

This subsection analyzes how the joint use of multiple feature extractors in the multi-feature setting affects the performance of SGC across different image classification datasets.
In this setting, both the graph structure and the node features are constructed from multiple backbone descriptors, which are combined through rank aggregation and feature fusion mechanisms.

For the multi-feature aggregation setting, we adopted a slightly larger neighborhood size, \(K_g = K_m = 50\), in order to better capture the complementary information provided by multiple heterogeneous descriptors. Since each feature extractor induces a distinct similarity structure, increasing \(K\) allows the re-ranking and manifold learning procedures to integrate a broader and more stable set of cross-feature relationships, without introducing noise or altering the local nature of the neighborhood construction.

\subsubsection{Parameter Analysis for Feature Selection}

URelief is used in the multi-feature setting by enabling the fusion of heterogeneous descriptors into a single representation.
Since each backbone produces features with different biases and levels of redundancy, a naive concatenation can amplify irrelevant dimensions and dilute discriminative ones.
In contrast, URelief ranks dimensions according to their ability to preserve local neighborhood consistency, selecting the most informative components across extractors while discarding noisy or redundant ones.

\begin{figure}[ht!]
\centering
\includegraphics[width=.48\textwidth]{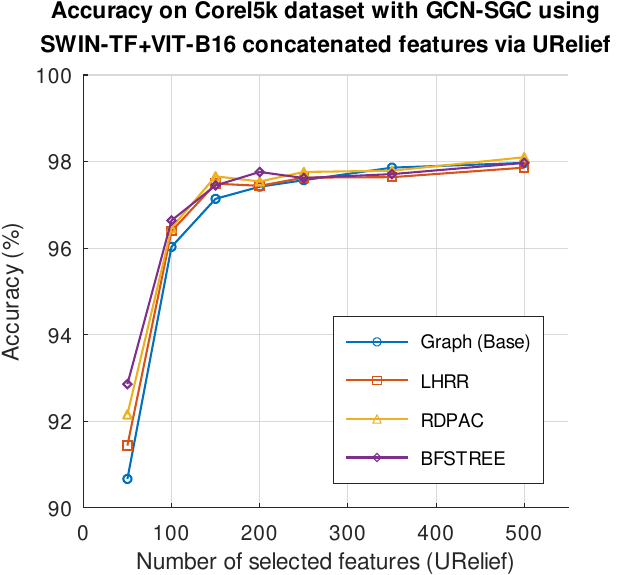}
\caption{Effect of the number of features selected by URelief on the Corel5k dataset using SGC with concatenated SWIN-TF and VIT-B16 features.}
\label{fig:corel5k_sgc_ablation}

\end{figure}

Therefore, we conducted an evaluation to determine the feature dimensionality for URelief, which is used to select the number of features from each extractor before concatenation. Figure~\ref{fig:corel5k_sgc_ablation} illustrates this analysis. It shows that performance quickly reaches a plateau at around 200 features without and with any of the manifold learning approaches.
Consequently, we adopt a fixed dimensionality of 200 selected features for URelief across all datasets and experimental settings.

\subsubsection{Multi-Feature Results}

Tables~\ref{tab:rankagg_flowers}, \ref{tab:rankagg_corel5k}, and \ref{tab:rankagg_cub200} present the SGC results when using reciprocal kNN graphs combined with rank aggregation, considering both single and multi feature descriptors (built from the three best backbones selected for each dataset). Overall, the three tables show a consistent pattern: (1) using SWIN-TF as the graph descriptor yields the most stable and highest accuracies across blocks, indicating that its similarity structure provides a strong manifold for propagation; (2) manifold learning usually further improves the aggregated setting, with RDPAC being the most effective option in the majority of cases; and (3) although some cross graph choices can degrade performance, multi-feature descriptors remain competitive and frequently match or surpass the best single-feature baselines, confirming that the proposed aggregation can exploit complementary rankings without losing the benefits of a strong graph backbone.

For the Flowers dataset (Table~\ref{tab:rankagg_flowers}), the best result is achieved by combining SWIN-TF+VIT-B16+CNN-ResNet features while keeping the graph as SWIN-TF and applying RDPAC (99.85\%), which slightly improves over the best single-feature configuration. This table also highlights that, when the graph is replaced by weaker alternatives (e.g., CNN-ResNet), performance drops sharply even with aggregated features, reinforcing that graph quality is the dominant factor in this dataset.

Regarding Corel5k (Table~\ref{tab:rankagg_corel5k}), the gains from rank aggregation and manifold learning are more evident, with the overall best score obtained using VIT-B16 features, a SWIN-TF graph, and RDPAC (98.52\%).

In the case of CUB200 (Table~\ref{tab:rankagg_cub200}), the dataset is harder due to higher intra-class and inter-class variability, and the results clearly emphasize the importance of both cross-combinations and manifold learning. The highest accuracy is obtained with VIT-B16 features using SWIN-TF as graph and RDPAC (83.33\%), outperforming the corresponding single-feature baselines and showing that the aggregated framework better captures fine-grained similarities when a transformer-based graph is adopted. Across all blocks, configurations relying on CNN-ResNet graphs remain limited, whereas using SWIN-TF as graph consistently lifts performance, with RDPAC yielding the largest improvements among the manifold learning options.

As a consolidated view of the results presented in this subsection (Tables~\ref{tab:rankagg_flowers},~\ref{tab:rankagg_corel5k}, and~\ref{tab:rankagg_cub200}), Table~\ref{tab:best_single_vs_combination_summary} summarizes and compares the best single-feature baselines without manifold learning (abbreviated as M.L.) against the best multi-feature SGC configurations for each dataset.  This table highlights the accuracy gains achieved by combining heterogeneous feature extractors and graph constructions, providing a compact overview of the benefits of feature fusion across datasets. Notice that the gains are higher for the largest dataset (CUB200). For Flowers, the gains are modest, since the baseline is already close to 100\%.
A statistical analysis was also conducted for each row of the table to assess the significance of the observed improvements using two tests: the Wilcoxon signed-rank test $(p < 0.05)$ and Cohen’s $d$. Overall, the results indicate not only statistically significant improvements, but also effect sizes that are practically meaningful across all three datasets.

\section{Conclusion}
\label{sec:conclusion}

This work investigated semi-supervised image classification with GNNs under limited-label scenarios by proposing a framework for combinations in both single-feature and multi-feature settings. In the single-feature scenario, including cross-combinations where node features and graph structures are derived from different backbone extractors, the results consistently show that using the same descriptor for both inputs is rarely optimal. Instead, heterogeneous combinations significantly improve classification accuracy, particularly when a strong backbone is used to construct the graph while complementary descriptors define the node features. These gains are further amplified by manifold learning techniques, which refine neighborhood relationships and improve message propagation, with RDPAC emerging as the most effective approach in most cases. In the multi-feature scenario, rank aggregation enables the integration of multiple ranked lists into a unified similarity structure, while feature fusion based on URelief and concatenation produces compact and informative node representations. This joint aggregation strategy is competitive and frequently surpasses the best single-descriptor baselines, especially on more challenging datasets, revealing that complementary information can be effectively combined without sacrificing the advantages of high-quality transformer-based representations.

For future work, we plan to investigate new ways to evaluate feature combinations. This includes using alternative graph constructions to capture different relationships, aggregating them through pooling mechanisms, experimenting with different dimensionality-reduction methods for rank aggregation, and further evaluating the impact of parameters such as neighborhood size. In addition, we intend to assess the proposed approaches across multiple datasets and other modalities (e.g., text, audio).

\section*{Declarations}

\begin{contributions}
Lucas Valem, Mohand Allili, and Daniel Pedronette contributed to the conception of this study. Marina Gapski conducted the preliminary experiments. Vinicius Kawai, Gustavo Leticio, and Lucas Valem conducted the definitive experiments. Marina Gapski, Vinicius Kawai, Gustavo Leticio, and Lucas Valem are the main contributors and writers of this manuscript. Marina Gapski, Vinicius Kawai, Gustavo Leticio, and Lucas Valem wrote the final text of the manuscript. 
All authors contributed to the final revision of the experiments and the manuscript. All authors read and approved the final manuscript. 
\end{contributions} 

\begin{interests}
The authors declare that they have no competing interests. 
\end{interests}

\begin{acknowledgements}
The authors gratefully acknowledge the financial support provided by the S\~{a}o Paulo Research Foundation -- FAPESP (grants \#2025/10602-5, \#2023/12736-3, and \#2018/15597-6), the Brazilian National Council for Scientific and Technological Development -- CNPq (grants \#313193/2023-1 and \#422667/2021-8), the Coordena\c{c}\~{a}o de Aperfei\c{c}oamento de Pessoal de N\'{i}vel Superior -- Brasil (CAPES, Finance Code 001), the University of S\~{a}o Paulo (PRPI Ordinance No.~1032, ``Apoio aos Novos Docentes''), and Petrobras (grant \#2023/00095-3).
\end{acknowledgements}

\begin{materials}
The source code developed in this work is available at \url{https://github.com/marinachagasbg/gnn-multifeature-fusion}. The datasets used in this study are publicly available in their respective sources.
\end{materials}

\bibliographystyle{plainnat}

\begin{table*}[ht!]
\caption{Accuracy (\%) of single-feature, cross-combination experiments on Flowers~\cite{Flowers} dataset for SGC model. The highest accuracy is highlighted in blue.}
\label{tab:results-sgc-flowers}
\centering
\resizebox{.95\textwidth}{!}{
\begin{tabular}{|c|c|c|c|c|c|}
\hline
 Features   & Graph      &  No Manifold Learning   & LHRR           & RDPAC          & BFSTREE        \\
 \hline
 \multirow{7}{*}{CNN-ResNet} & CNN-ResNet & 79.71 $\pm$ 0.0535 & 84.40 $\pm$ 0.0572 & 84.10 $\pm$ 0.0751 & 83.34 $\pm$ 0.0558 \\
  & CNN-DPNet  & 78.19 $\pm$ 0.0686 & 82.34 $\pm$ 0.0725 & 82.20 $\pm$ 0.0491 & 82.06 $\pm$ 0.0610 \\
  & CNN-SENet  & 74.24 $\pm$ 0.0542 & 75.68 $\pm$ 0.0850 & 76.42 $\pm$ 0.0811 & 77.32 $\pm$ 0.0122 \\
  & T2T-VIT24  & 72.66 $\pm$ 0.1273 & 73.43 $\pm$ 0.0878 & 75.43 $\pm$ 0.0327 & 73.84 $\pm$ 0.0873 \\
  & VIT-B16    & 93.21 $\pm$ 0.0545 & 95.64 $\pm$ 0.0348 & 97.24 $\pm$ 0.0217 & 96.49 $\pm$ 0.0411 \\
  & SWIN-TF    & \textbf{95.19 $\pm$ 0.0515} & \textbf{98.67 $\pm$ 0.0415} & \textbf{99.71 $\pm$ 0.0033} & \textbf{99.47 $\pm$ 0.0052} \\
  & ConvNeXt   & 73.90 $\pm$ 0.0666 & 74.71 $\pm$ 0.0523 & 77.23 $\pm$ 0.0636 & 76.82 $\pm$ 0.0504 \\
 \hline
 \multirow{7}{*}{CNN-DPNet}  & CNN-ResNet & 78.11 $\pm$ 0.0354 & 83.75 $\pm$ 0.0608 & 83.42 $\pm$ 0.0367 & 82.73 $\pm$ 0.0364 \\
   & CNN-DPNet  & 77.12 $\pm$ 0.0729 & 80.32 $\pm$ 0.0877 & 81.09 $\pm$ 0.0552 & 81.46 $\pm$ 0.1221 \\
   & CNN-SENet  & 72.97 $\pm$ 0.1045 & 74.41 $\pm$ 0.0784 & 75.20 $\pm$ 0.0715 & 75.75 $\pm$ 0.0805 \\
   & T2T-VIT24  & 70.33 $\pm$ 0.1484 & 71.96 $\pm$ 0.0290 & 74.23 $\pm$ 0.1061 & 72.84 $\pm$ 0.0489 \\
   & VIT-B16    & 91.89 $\pm$ 0.0544 & 95.36 $\pm$ 0.0557 & 96.71 $\pm$ 0.0286 & 96.21 $\pm$ 0.0317 \\
   & SWIN-TF    & \textbf{93.59 $\pm$ 0.0517} & \textbf{97.96 $\pm$ 0.0478} & \textbf{99.70 $\pm$ 0.0061} & \textbf{99.45 $\pm$ 0.0033} \\
   & ConvNeXt   & 72.75 $\pm$ 0.0599 & 73.47 $\pm$ 0.1430 & 76.39 $\pm$ 0.1073 & 75.92 $\pm$ 0.0480 \\
 \hline
 \multirow{7}{*}{CNN-SENet}  & CNN-ResNet & 79.36 $\pm$ 0.0965 & 83.96 $\pm$ 0.0574 & 83.93 $\pm$ 0.0675 & 83.10 $\pm$ 0.0339  \\
   & CNN-DPNet  & 78.31 $\pm$ 0.0800 & 81.14 $\pm$ 0.1008 & 81.96 $\pm$ 0.0328 & 82.02 $\pm$ 0.0890 \\
   & CNN-SENet  & 72.98 $\pm$ 0.0796 & 74.00 $\pm$ 0.0810 & 75.06 $\pm$ 0.0360 & 75.13 $\pm$ 0.0329 \\
   & T2T-VIT24  & 70.60 $\pm$ 0.1037 & 72.28 $\pm$ 0.1096 & 75.45 $\pm$ 0.0704 & 72.89 $\pm$ 0.0680 \\
   & VIT-B16    & 92.34 $\pm$ 0.0415 & 95.26 $\pm$ 0.0229 & 96.97 $\pm$ 0.0122 & 96.33 $\pm$ 0.0122 \\
   & SWIN-TF    & \textbf{95.18 $\pm$ 0.0710} & \textbf{98.52 $\pm$ 0.0228} & \textcolor{blue}{\textbf{99.73 $\pm$ 0.0033}} & \textbf{99.41 $\pm$ 0.0120} \\
   & ConvNeXt   & 72.58 $\pm$ 0.1402 & 73.75 $\pm$ 0.0562 & 76.97 $\pm$ 0.0481 & 76.32 $\pm$ 0.0573 \\
 \hline
 \multirow{7}{*}{T2T-VIT24}  & CNN-ResNet & 79.32 $\pm$ 0.0907 & 83.80 $\pm$ 0.0486 & 84.13 $\pm$ 0.0370 & 83.34 $\pm$ 0.0638 \\
   & CNN-DPNet  & 77.86 $\pm$ 0.1106 & 80.99 $\pm$ 0.1525 & 81.91 $\pm$ 0.0635 & 81.86 $\pm$ 0.0421 \\
   & CNN-SENet  & 72.93 $\pm$ 0.1181 & 75.04 $\pm$ 0.0586 & 76.57 $\pm$ 0.0551 & 76.60 $\pm$ 0.0636  \\
   & T2T-VIT24  & 70.01 $\pm$ 0.0875 & 72.24 $\pm$ 0.0744 & 75.18 $\pm$ 0.0464 & 72.81 $\pm$ 0.0396 \\
   & VIT-B16    & 91.82 $\pm$ 0.1026 & 95.06 $\pm$ 0.0452 & 96.41 $\pm$ 0.0565 & 95.94 $\pm$ 0.0248 \\
   & SWIN-TF    & \textbf{94.54 $\pm$ 0.0508} & \textbf{98.32 $\pm$ 0.0116} & \textbf{99.69 $\pm$ 0.0052} & \textbf{99.42 $\pm$ 0.0065} \\
   & ConvNeXt   & 73.13 $\pm$ 0.0904 & 74.00 $\pm$ 0.0446 & 77.02 $\pm$ 0.0387 & 76.51 $\pm$ 0.0353 \\
 \hline
 \multirow{7}{*}{VIT-B16}    & CNN-ResNet & 83.72 $\pm$ 0.0946 & 88.41 $\pm$ 0.0701 & 86.89 $\pm$ 0.0203 & 86.57 $\pm$ 0.0460 \\
     & CNN-DPNet  & 80.50 $\pm$ 0.0467 & 85.54 $\pm$ 0.1778 & 84.96 $\pm$ 0.0543 & 85.32 $\pm$ 0.1092 \\
     & CNN-SENet  & 77.38 $\pm$ 0.1694 & 80.02 $\pm$ 0.0543 & 79.31 $\pm$ 0.1119 & 81.63 $\pm$ 0.0930 \\
     & T2T-VIT24  & 75.64 $\pm$ 0.0507 & 77.33 $\pm$ 0.1030 & 78.46 $\pm$ 0.0873 & 76.87 $\pm$ 0.0742 \\
     & VIT-B16    & 92.85 $\pm$ 0.0308 & 95.89 $\pm$ 0.0378 & 96.95 $\pm$ 0.0228 & 96.39 $\pm$ 0.0360 \\
     & SWIN-TF    & \textbf{97.33 $\pm$ 0.0339} & \textbf{99.20 $\pm$ 0.0116} & \textbf{99.49 $\pm$ 0.0000} & \textbf{99.49 $\pm$ 0.0000} \\
     & ConvNeXt   & 77.56 $\pm$ 0.0942 & 78.22 $\pm$ 0.0698 & 79.45 $\pm$ 0.0771 & 78.90 $\pm$ 0.0617 \\
 \hline
 \multirow{7}{*}{SWIN-TF}    & CNN-ResNet & 84.03 $\pm$ 0.0623 & 86.46 $\pm$ 0.0419 & 85.69 $\pm$ 0.0335 & 85.36 $\pm$ 0.0152 \\
     & CNN-DPNet  & 82.52 $\pm$ 0.0352 & 83.94 $\pm$ 0.1151 & 83.76 $\pm$ 0.0152 & 83.78 $\pm$ 0.0152 \\
     & CNN-SENet  & 77.51 $\pm$ 0.0469 & 77.64 $\pm$ 0.0449 & 77.92 $\pm$ 0.0425 & 79.34 $\pm$ 0.0411 \\
     & T2T-VIT24  & 76.96 $\pm$ 0.1055 & 75.15 $\pm$ 0.0320 & 76.49 $\pm$ 0.0228 & 75.77 $\pm$ 0.0524 \\
     & VIT-B16    & 94.63 $\pm$ 0.0290 & 95.93 $\pm$ 0.0152 & 97.25 $\pm$ 0.0152 & 96.61 $\pm$ 0.0089 \\
     & SWIN-TF    & \textbf{97.06 $\pm$ 0.0331} & \textbf{99.02 $\pm$ 0.0116} & \textbf{99.58 $\pm$ 0.0052} & \textbf{99.44 $\pm$ 0.0033} \\
     & ConvNeXt   & 77.38 $\pm$ 0.0711 & 76.53 $\pm$ 0.0314 & 78.19 $\pm$ 0.0363 & 78.40 $\pm$ 0.0348  \\
 \hline
 \multirow{7}{*}{ConvNeXt}  & CNN-ResNet & 79.21 $\pm$ 0.0333 & 84.68 $\pm$ 0.0545 & 84.72 $\pm$ 0.0245 & 83.30 $\pm$ 0.0223 \\
    & CNN-DPNet  & 77.66 $\pm$ 0.0261 & 82.25 $\pm$ 0.0290 & 82.76 $\pm$ 0.0222 & 82.54 $\pm$ 0.0142 \\
    & CNN-SENet  & 73.82 $\pm$ 0.0595 & 76.07 $\pm$ 0.0222 & 76.70 $\pm$ 0.0329 & 77.43 $\pm$ 0.0248 \\
    & T2T-VIT24  & 71.52 $\pm$ 0.0525 & 73.86 $\pm$ 0.0487 & 76.19 $\pm$ 0.0513 & 74.14 $\pm$ 0.0111 \\
    & VIT-B16    & 92.67 $\pm$ 0.0133 & 95.48 $\pm$ 0.0083 & 97.37 $\pm$ 0.0089 & 96.54 $\pm$ 0.0083 \\
    & SWIN-TF    & \textbf{94.37 $\pm$ 0.0543} & \textbf{98.40 $\pm$ 0.0167} & \textbf{99.66 $\pm$ 0.0052} & \textbf{99.50 $\pm$ 0.0000} \\
    & ConvNeXt   & 73.32 $\pm$ 0.0478 & 74.75 $\pm$ 0.0364 & 76.88 $\pm$ 0.0176 & 76.49 $\pm$ 0.0429 \\
\hline
\end{tabular}
}
\end{table*}

\begin{table*}[ht!]
\caption{Accuracy (\%) of single-feature, cross-combination experiments on Corel5k~\cite{corel5k} dataset for SGC model. The highest accuracy is highlighted in blue.}
\label{tab:results-sgc-corel5k}
\centering
\resizebox{.85\textwidth}{!}{
\begin{tabular}{|c|c|c|c|c|c|}
\hline
 Features   & Graph      & No Manifold Learning & LHRR           & RDPAC          & BFSTREE        \\
 \hline
 \multirow{8}{*}{CNN-ResNet} & CNN-ResNet & 89.59 $\pm$ 0.0541 & 91.22 $\pm$ 0.0203 & 91.47 $\pm$ 0.0329 & 91.95 $\pm$ 0.0212 \\
  & CNN-DPNet  & 88.17 $\pm$ 0.0496 & 89.88 $\pm$ 0.0618 & 90.05 $\pm$ 0.0360 & 90.10 $\pm$ 0.0299 \\
  & CNN-SENet  & 90.18 $\pm$ 0.0624 & 90.39 $\pm$ 0.0601 & 90.84 $\pm$ 0.0260 & 91.45 $\pm$ 0.0242 \\
  & T2T-VIT24  & 90.22 $\pm$ 0.0348 & 91.37 $\pm$ 0.0792 & 91.97 $\pm$ 0.0421 & 92.10 $\pm$ 0.0262 \\
  & VIT-B16    & 94.16 $\pm$ 0.0187 & 94.91 $\pm$ 0.0241 & 94.46 $\pm$ 0.0342 & 94.70 $\pm$ 0.0195 \\
  & SWIN-TF    & \textbf{95.24 $\pm$ 0.0366} & \textbf{96.10 $\pm$ 0.0685} & \textbf{97.73 $\pm$ 0.0346} & \textbf{97.44 $\pm$ 0.0512} \\
  & DINOv2     & 93.76 $\pm$ 0.0385 & 95.40 $\pm$ 0.0408 & 95.76 $\pm$ 0.0382 & 95.72 $\pm$ 0.0511 \\
  & ConvNeXt   & 89.42 $\pm$ 0.0395 & 91.06 $\pm$ 0.0595 & 91.79 $\pm$ 0.0358 & 92.00 $\pm$ 0.0210 \\
 \hline
 \multirow{8}{*}{CNN-DPNet}  & CNN-ResNet & 89.73 $\pm$ 0.0386 & 91.24 $\pm$ 0.0123 & 91.21 $\pm$ 0.0504 & 92.01 $\pm$ 0.0235 \\
   & CNN-DPNet  & 86.77 $\pm$ 0.1132 & 88.84 $\pm$ 0.1015 & 88.92 $\pm$ 0.0989 & 89.28 $\pm$ 0.0246 \\
   & CNN-SENet  & 89.21 $\pm$ 0.0279 & 89.77 $\pm$ 0.0380 & 90.34 $\pm$ 0.0257 & 91.04 $\pm$ 0.0376 \\
   & T2T-VIT24  & 88.70 $\pm$ 0.0521 & 90.65 $\pm$ 0.0474 & 91.61 $\pm$ 0.0422 & 91.76 $\pm$ 0.0471 \\
   & VIT-B16    & 93.30 $\pm$ 0.0747 & 94.14 $\pm$ 0.0609 & 94.03 $\pm$ 0.0651 & 94.15 $\pm$ 0.1241 \\
   & SWIN-TF    & \textbf{94.57 $\pm$ 0.0480} & \textbf{96.09 $\pm$ 0.0555} & \textbf{97.17 $\pm$ 0.0450} & \textbf{96.94 $\pm$ 0.0815} \\
   & DINOv2     & 93.54 $\pm$ 0.1151 & 95.08 $\pm$ 0.0426 & 95.55 $\pm$ 0.0578 & 95.37 $\pm$ 0.0431 \\
   & ConvNeXt   & 88.28 $\pm$ 0.0760 & 90.46 $\pm$ 0.0351 & 91.43 $\pm$ 0.0575 & 91.63 $\pm$ 0.0402 \\
 \hline
 \multirow{8}{*}{CNN-SENet}  & CNN-ResNet & 90.15 $\pm$ 0.0346 & 91.62 $\pm$ 0.0309 & 91.50 $\pm$ 0.0244 & 92.15 $\pm$ 0.0202 \\
   & CNN-DPNet  & 88.35 $\pm$ 0.0350 & 89.59 $\pm$ 0.0387 & 89.83 $\pm$ 0.0693 & 89.83 $\pm$ 0.0370 \\
   & CNN-SENet  & 89.84 $\pm$ 0.0421 & 89.92 $\pm$ 0.0574 & 90.66 $\pm$ 0.0451 & 91.41 $\pm$ 0.0137 \\
   & T2T-VIT24  & 89.91 $\pm$ 0.0258 & 90.84 $\pm$ 0.0465 & 91.96 $\pm$ 0.0476 & 92.06 $\pm$ 0.0500 \\
   & VIT-B16    & 93.94 $\pm$ 0.0402 & 94.77 $\pm$ 0.0538 & 94.25 $\pm$ 0.0225 & 94.59 $\pm$ 0.0106 \\
   & SWIN-TF    & \textbf{95.31 $\pm$ 0.0165} & \textbf{96.32 $\pm$ 0.0209} & \textbf{97.63 $\pm$ 0.0342} & \textbf{97.52 $\pm$ 0.0187} \\
   & DINOv2     & 93.84 $\pm$ 0.0275 & 95.51 $\pm$ 0.0490 & 95.59 $\pm$ 0.0381 & 95.65 $\pm$ 0.0364 \\
   & ConvNeXt   & 89.36 $\pm$ 0.0207 & 90.82 $\pm$ 0.0377 & 91.24 $\pm$ 0.0217 & 91.84 $\pm$ 0.0324 \\
 \hline
 \multirow{8}{*}{T2T-VIT24}  & CNN-ResNet & 90.32 $\pm$ 0.0553 & 91.73 $\pm$ 0.0582 & 91.61 $\pm$ 0.0435 & 92.38 $\pm$ 0.0363 \\
   & CNN-DPNet  & 88.25 $\pm$ 0.0326 & 89.66 $\pm$ 0.0386 & 89.94 $\pm$ 0.0483 & 89.99 $\pm$ 0.0353 \\
   & CNN-SENet  & 90.12 $\pm$ 0.0395 & 90.30 $\pm$ 0.0546 & 91.19 $\pm$ 0.0498 & 91.55 $\pm$ 0.0574 \\
   & T2T-VIT24  & 88.91 $\pm$ 0.0566 & 90.55 $\pm$ 0.0336 & 91.79 $\pm$ 0.0791 & 91.70 $\pm$ 0.0601 \\
   & VIT-B16    & 93.55 $\pm$ 0.0540 & 94.43 $\pm$ 0.0589 & 94.02 $\pm$ 0.0291 & 94.45 $\pm$ 0.0392 \\
   & SWIN-TF    & \textbf{95.34 $\pm$ 0.0447} & \textbf{96.16 $\pm$ 0.0362} & \textbf{97.33 $\pm$ 0.0644} & \textbf{97.26 $\pm$ 0.0591} \\
   & DINOv2     & 93.94 $\pm$ 0.0286 & 95.44 $\pm$ 0.1092 & 95.36 $\pm$ 0.0296 & 95.65 $\pm$ 0.0643 \\
   & ConvNeXt   & 89.29 $\pm$ 0.0421 & 91.03 $\pm$ 0.1197 & 91.52 $\pm$ 0.0478 & 91.70 $\pm$ 0.0499 \\
 \hline
 \multirow{8}{*}{VIT-B16}    & CNN-ResNet & 91.88 $\pm$ 0.0428 & 93.48 $\pm$ 0.0646 & 93.35 $\pm$ 0.0418 & 93.70 $\pm$ 0.0357 \\
     & CNN-DPNet  & 90.10 $\pm$ 0.0463 & 92.29 $\pm$ 0.0593 & 92.30 $\pm$ 0.0292 & 92.34 $\pm$ 0.0337 \\
     & CNN-SENet  & 91.99 $\pm$ 0.0347 & 92.62 $\pm$ 0.0314 & 93.00 $\pm$ 0.0546 & 93.39 $\pm$ 0.0145 \\
     & T2T-VIT24  & 91.35 $\pm$ 0.0447 & 92.83 $\pm$ 0.0609 & 93.22 $\pm$ 0.0250 & 93.44 $\pm$ 0.0424 \\
     & VIT-B16    & 93.40 $\pm$ 0.0301 & 95.22 $\pm$ 0.0267 & 94.76 $\pm$ 0.0469 & 95.29 $\pm$ 0.0372 \\
     & SWIN-TF    & \textbf{96.39 $\pm$ 0.0409} & \textbf{97.24 $\pm$ 0.0827} & \textcolor{blue}{\textbf{98.53 $\pm$ 0.0090}} & \textbf{98.38 $\pm$ 0.0141} \\
     & DINOv2     & 94.68 $\pm$ 0.0355 & 96.54 $\pm$ 0.0204 & 96.57 $\pm$ 0.0308 & 96.58 $\pm$ 0.0346 \\
     & ConvNeXt   & 90.57 $\pm$ 0.0470 & 92.74 $\pm$ 0.1172 & 93.06 $\pm$ 0.0484 & 93.70 $\pm$ 0.0374 \\
 \hline
 \multirow{8}{*}{SWIN-TF}    & CNN-ResNet & 91.29 $\pm$ 0.0266 & 92.34 $\pm$ 0.0168 & 92.37 $\pm$ 0.0239 & 92.80 $\pm$ 0.0099 \\
     & CNN-DPNet  & 89.69 $\pm$ 0.0236 & 90.50 $\pm$ 0.0487 & 90.79 $\pm$ 0.0334 & 90.67 $\pm$ 0.0259 \\
     & CNN-SENet  & 91.33 $\pm$ 0.0337 & 91.07 $\pm$ 0.0320 & 91.70 $\pm$ 0.0434 & 92.07 $\pm$ 0.0722 \\
     & T2T-VIT24  & 91.45 $\pm$ 0.0212 & 91.61 $\pm$ 0.0306 & 92.06 $\pm$ 0.0191 & 92.71 $\pm$ 0.0162 \\
     & VIT-B16    & 94.41 $\pm$ 0.0472 & 95.21 $\pm$ 0.0633 & 94.62 $\pm$ 0.0338 & 95.17 $\pm$ 0.0671 \\
     & SWIN-TF    & \textbf{95.89 $\pm$ 0.0276} & \textbf{96.60 $\pm$ 0.0606} & \textbf{98.23 $\pm$ 0.0165} & \textbf{97.86 $\pm$ 0.0542} \\
     & DINOv2     & 94.46 $\pm$ 0.0258 & 95.87 $\pm$ 0.0116 & 95.94 $\pm$ 0.0103 & 95.87 $\pm$ 0.0283 \\
     & ConvNeXt   & 90.81 $\pm$ 0.0424 & 91.51 $\pm$ 0.0380 & 92.22 $\pm$ 0.0359 & 92.71 $\pm$ 0.0354 \\
 \hline
 \multirow{8}{*}{DINOv2}     & CNN-ResNet & 90.83 $\pm$ 0.0668 & 91.83 $\pm$ 0.0512 & 91.78 $\pm$ 0.0614 & 92.41 $\pm$ 0.0367 \\
      & CNN-DPNet  & 89.14 $\pm$ 0.0531 & 90.21 $\pm$ 0.0152 & 90.10 $\pm$ 0.1200 & 90.17 $\pm$ 0.0949 \\
      & CNN-SENet  & 91.11 $\pm$ 0.0412 & 90.34 $\pm$ 0.0451 & 91.27 $\pm$ 0.0813 & 91.56 $\pm$ 0.0418 \\
      & T2T-VIT24  & 90.91 $\pm$ 0.0152 & 90.83 $\pm$ 0.1007 & 91.46 $\pm$ 0.1288 & 92.25 $\pm$ 0.0244 \\
      & VIT-B16    & 93.94 $\pm$ 0.0572 & 94.56 $\pm$ 0.0975 & 94.07 $\pm$ 0.1105 & 94.01 $\pm$ 0.1126 \\
      & SWIN-TF    & \textbf{95.68 $\pm$ 0.0547} & \textbf{96.17 $\pm$ 0.0924} & \textbf{97.50 $\pm$ 0.0579} & \textbf{97.47 $\pm$ 0.0421} \\
      & DINOv2     & 93.37 $\pm$ 0.0550 & 95.11 $\pm$ 0.0971 & 94.86 $\pm$ 0.0557 & 94.99 $\pm$ 0.0797 \\
      & ConvNeXt   & 90.47 $\pm$ 0.0576 & 91.14 $\pm$ 0.0647 & 91.66 $\pm$ 0.0749 & 92.04 $\pm$ 0.0497 \\
 \hline
 \multirow{8}{*}{ConvNeXt}   & CNN-ResNet & 91.29 $\pm$ 0.0226 & 92.83 $\pm$ 0.0470 & 92.51 $\pm$ 0.0051 & 93.23 $\pm$ 0.0173 \\
    & CNN-DPNet  & 89.20 $\pm$ 0.0187 & 91.30 $\pm$ 0.0128 & 91.27 $\pm$ 0.0230 & 91.52 $\pm$ 0.0438 \\
    & CNN-SENet  & 90.95 $\pm$ 0.0177 & 91.47 $\pm$ 0.0340 & 92.00 $\pm$ 0.0049 & 92.68 $\pm$ 0.0290 \\
    & T2T-VIT24  & 90.59 $\pm$ 0.0351 & 91.75 $\pm$ 0.0346 & 92.61 $\pm$ 0.0094 & 93.00 $\pm$ 0.0065 \\
    & VIT-B16    & 94.36 $\pm$ 0.0081 & 95.00 $\pm$ 0.0269 & 94.58 $\pm$ 0.0413 & 94.99 $\pm$ 0.0056 \\
    & SWIN-TF    & \textbf{95.70 $\pm$ 0.0160} & \textbf{96.79 $\pm$ 0.0657} & \textbf{97.51 $\pm$ 0.0027} & \textbf{97.64 $\pm$ 0.0219} \\
    & DINOv2     & 94.80 $\pm$ 0.0180 & 95.57 $\pm$ 0.0502 & 95.66 $\pm$ 0.0104 & 95.97 $\pm$ 0.0127 \\
    & ConvNeXt   & 89.81 $\pm$ 0.0076 & 91.29 $\pm$ 0.0319 & 91.92 $\pm$ 0.0191 & 92.72 $\pm$ 0.0438 \\
\hline
\end{tabular}
}
\end{table*}

\begin{table*}[ht!]
\caption{Accuracy (\%) of single-feature, cross-combination experiments on  CUB200~\cite{cub200} dataset for SGC model. The highest accuracy is highlighted in blue.}
\label{tab:results-sgc-cub200}
\centering
\resizebox{.95\textwidth}{!}{
\begin{tabular}{|c|c|c|c|c|c|}
\hline
 Features   & Graph      & No Manifold Learning  & LHRR           & RDPAC          & BFSTREE        \\
\hline
 \multirow{7}{*}{CNN-ResNet} & CNN-ResNet & 47.03 $\pm$ 0.0203 & 51.53 $\pm$ 0.0317 & 51.81 $\pm$ 0.0340 & 51.67 $\pm$ 0.0331 \\
  & CNN-DPNet  & 48.15 $\pm$ 0.0508 & 52.19 $\pm$ 0.0216 & 53.35 $\pm$ 0.0324 & 53.00 $\pm$ 0.0430   \\
  & CNN-SENet  & 37.85 $\pm$ 0.0425 & 37.37 $\pm$ 0.0382 & 38.85 $\pm$ 0.0249 & 39.38 $\pm$ 0.0547 \\
  & T2T-VIT24  & 41.09 $\pm$ 0.0281 & 40.02 $\pm$ 0.0368 & 42.08 $\pm$ 0.0260 & 42.51 $\pm$ 0.0700   \\
  & VIT-B16    & 74.26 $\pm$ 0.0178 & 77.57 $\pm$ 0.0472 & 78.80 $\pm$ 0.0108 & 78.16 $\pm$ 0.0213 \\
  & SWIN-TF    & \textbf{75.74 $\pm$ 0.0377} & \textbf{79.47 $\pm$ 0.0188} & \textcolor{blue}{\textbf{82.25 $\pm$ 0.0227}} & \textbf{80.38 $\pm$ 0.0231} \\
  & ConvNeXt   & 42.60 $\pm$ 0.0273  & 39.67 $\pm$ 0.0821 & 42.00 $\pm$ 0.0489  & 42.52 $\pm$ 0.0305 \\
  \hline
 \multirow{7}{*}{CNN-DPNet}  & CNN-ResNet & 40.51 $\pm$ 0.4443 & 50.38 $\pm$ 0.0930  & 49.70 $\pm$ 0.0790   & 49.52 $\pm$ 0.0756 \\
   & CNN-DPNet  & 45.40 $\pm$ 0.1891 & 50.92 $\pm$ 0.0622 & 51.15 $\pm$ 0.0532 & 51.00 $\pm$ 0.0953  \\
   & CNN-SENet  & 30.28 $\pm$ 0.3589 & 36.39 $\pm$ 0.1325 & 36.94 $\pm$ 0.1117 & 37.50 $\pm$ 0.0576  \\
   & T2T-VIT24  & 33.12 $\pm$ 0.3675 & 37.91 $\pm$ 0.4000 & 40.44 $\pm$ 0.1447 & 40.84 $\pm$ 0.1117 \\
   & VIT-B16    & 71.72 $\pm$ 0.1885 & 76.81 $\pm$ 0.0979 & 77.91 $\pm$ 0.0485 & 76.87 $\pm$ 0.0653 \\
   & SWIN-TF    & \textbf{73.10 $\pm$ 0.2510  } & \textbf{78.85 $\pm$ 0.2415} & \textbf{81.58 $\pm$ 0.0663} & \textbf{79.76 $\pm$ 0.0488} \\
   & ConvNeXt   & 36.34 $\pm$ 0.0961 & 38.69 $\pm$ 0.0622 & 40.20 $\pm$ 0.0705  & 41.10 $\pm$ 0.0938  \\
  \hline
 \multirow{7}{*}{CNN-SENet}  & CNN-ResNet & 44.98 $\pm$ 0.0679 & 50.52 $\pm$ 0.0431 & 51.39 $\pm$ 0.0197 & 51.11 $\pm$ 0.0346 \\
   & CNN-DPNet  & 46.74 $\pm$ 0.0352 & 50.69 $\pm$ 0.0213 & 52.33 $\pm$ 0.0448 & 51.93 $\pm$ 0.0147 \\
   & CNN-SENet  & 36.27 $\pm$ 0.0481 & 35.62 $\pm$ 0.0292 & 37.54 $\pm$ 0.0236 & 37.53 $\pm$ 0.0391 \\
   & T2T-VIT24  & 38.71 $\pm$ 0.0610 & 38.34 $\pm$ 0.0184 & 41.22 $\pm$ 0.0116 & 40.73 $\pm$ 0.0218 \\
   & VIT-B16    & 73.40 $\pm$ 0.0162 & 76.43 $\pm$ 0.0130 & 78.36 $\pm$ 0.0260 & 77.58 $\pm$ 0.0144 \\
   & SWIN-TF    & \textbf{74.75 $\pm$ 0.0272} & \textbf{77.97 $\pm$ 0.0255} & \textbf{82.02 $\pm$ 0.0140} & \textbf{79.77 $\pm$ 0.0128} \\
   & ConvNeXt   & 40.40 $\pm$ 0.0397  & 38.02 $\pm$ 0.0368 & 40.91 $\pm$ 0.0106 & 40.96 $\pm$ 0.0242 \\
  \hline
 \multirow{7}{*}{VIT-B16}    & CNN-ResNet & 46.46 $\pm$ 0.2205 & 52.29 $\pm$ 0.1122 & 51.55 $\pm$ 0.0670 & 51.89 $\pm$ 0.0616 \\
     & CNN-DPNet  & 48.40 $\pm$ 0.0859 & 53.28 $\pm$ 0.0309 & 53.07 $\pm$ 0.0330 & 53.45 $\pm$ 0.0879 \\
     & CNN-SENet  & 36.75 $\pm$ 0.0718 & 38.46 $\pm$ 0.0471 & 38.52 $\pm$ 0.0773 & 39.57 $\pm$ 0.1070  \\
     & T2T-VIT24  & 41.59 $\pm$ 0.1819 & 40.76 $\pm$ 0.1085 & 42.30 $\pm$ 0.1034 & 42.97 $\pm$ 0.0477 \\
     & VIT-B16    & 73.89 $\pm$ 0.0820 & 77.76 $\pm$ 0.0426 & 78.21 $\pm$ 0.0510 & 77.45 $\pm$ 0.0819 \\
     & SWIN-TF    & \textbf{77.85 $\pm$ 0.0576} & \textbf{80.53 $\pm$ 0.1050 } & \textbf{82.03 $\pm$ 0.0388} & \textbf{80.65 $\pm$ 0.0409} \\
     & ConvNeXt   & 42.80 $\pm$ 0.0871  & 40.42 $\pm$ 0.1058 & 41.71 $\pm$ 0.1208 & 43.01 $\pm$ 0.0512 \\
  \hline
 \multirow{7}{*}{SWIN-TF}    & CNN-ResNet & 50.59 $\pm$ 0.0121 & 51.76 $\pm$ 0.0217 & 53.06 $\pm$ 0.0291 & 53.38 $\pm$ 0.0221 \\
     & CNN-DPNet  & 51.59 $\pm$ 0.0143 & 52.11 $\pm$ 0.0427 & 53.89 $\pm$ 0.0136 & 54.08 $\pm$ 0.0122 \\
     & CNN-SENet  & 39.88 $\pm$ 0.0134 & 37.16 $\pm$ 0.0203 & 39.56 $\pm$ 0.0199 & 40.33 $\pm$ 0.0212 \\
     & T2T-VIT24  & 45.06 $\pm$ 0.0072 & 39.61 $\pm$ 0.024  & 42.71 $\pm$ 0.0266 & 42.81 $\pm$ 0.0196 \\
     & VIT-B16    & 75.48 $\pm$ 0.0097 & 77.08 $\pm$ 0.031  & 78.66 $\pm$ 0.0188 & 78.07 $\pm$ 0.0236 \\
     & SWIN-TF    & \textbf{77.51 $\pm$ 0.0181} & \textbf{78.92 $\pm$ 0.0054} & \textbf{81.86 $\pm$ 0.0232} & \textbf{79.75 $\pm$ 0.0168} \\
     & ConvNeXt   & 44.78 $\pm$ 0.0110  & 39.02 $\pm$ 0.0136 & 42.31 $\pm$ 0.0333 & 42.89 $\pm$ 0.0145 \\
  \hline
 \multirow{7}{*}{ConvNeXt}   & CNN-ResNet & 46.25 $\pm$ 0.0087 & 51.25 $\pm$ 0.0092 & 50.96 $\pm$ 0.0023 & 50.90 $\pm$ 0.0016  \\
    & CNN-DPNet  & 47.35 $\pm$ 0.0105 & 51.47 $\pm$ 0.0090 & 51.84 $\pm$ 0.0043 & 51.82 $\pm$ 0.0040  \\
    & CNN-SENet  & 37.47 $\pm$ 0.0190 & 37.31 $\pm$ 0.0073 & 37.87 $\pm$ 0.0020 & 38.29 $\pm$ 0.0019 \\
    & T2T-VIT24  & 40.50 $\pm$ 0.0104 & 39.64 $\pm$ 0.0054 & 41.36 $\pm$ 0.0054 & 41.53 $\pm$ 0.0021 \\
    & VIT-B16    & 73.54 $\pm$ 0.0062 & 76.81 $\pm$ 0.0048 & 78.17 $\pm$ 0.0023 & 77.23 $\pm$ 0.0015 \\
    & SWIN-TF    & \textbf{74.01 $\pm$ 0.0058} & \textbf{78.37 $\pm$ 0.0077} & \textbf{81.67 $\pm$ 0.0008} & \textbf{80.01 $\pm$ 0.0023} \\
    & ConvNeXt   & 41.81 $\pm$ 0.0073 & 38.95 $\pm$ 0.0040  & 40.09 $\pm$ 0.0026 & 40.75 $\pm$ 0.0018 \\
  \hline
 \multirow{7}{*}{T2T-VIT24}  & CNN-ResNet & 45.87 $\pm$ 0.0237 & 50.13 $\pm$ 0.0210 & 51.48 $\pm$ 0.0229 & 51.07 $\pm$ 0.0173 \\
   & CNN-DPNet  & 47.08 $\pm$ 0.0159 & 50.35 $\pm$ 0.0143 & 52.64 $\pm$ 0.0293 & 52.17 $\pm$ 0.0069 \\
   & CNN-SENet  & 36.01 $\pm$ 0.0144 & 35.73 $\pm$ 0.0117 & 38.36 $\pm$ 0.0104 & 38.39 $\pm$ 0.0178 \\
   & T2T-VIT24  & 39.12 $\pm$ 0.0245 & 37.47 $\pm$ 0.0146 & 40.27 $\pm$ 0.0156 & 39.82 $\pm$ 0.0134 \\
   & VIT-B16    & 73.54 $\pm$ 0.0144 & 75.98 $\pm$ 0.0191 & 78.35 $\pm$ 0.0105 & 77.55 $\pm$ 0.0141 \\
   & SWIN-TF    & \textbf{74.21 $\pm$ 0.0167} & \textbf{77.17 $\pm$ 0.0259} & \textbf{81.79 $\pm$ 0.0155} & \textbf{79.20 $\pm$ 0.0158 } \\
   & ConvNeXt   & 40.30 $\pm$ 0.0180 & 37.56 $\pm$ 0.0166 & 40.87 $\pm$ 0.0173 & 40.91 $\pm$ 0.0157 \\
\hline
\end{tabular}
}
\end{table*}

\begin{table*}[!ht]
\centering
\caption{Accuracy (\%) of \emph{best single} and \emph{best combination} results across datasets for \textbf{SGC} using \textbf{single-feature graphs} and \textbf{single-extractor features}. For each manifold learning method (LHRR, RDPAC, BFSTREE), we report: \textbf{Best Single}, defined as the highest mean accuracy among results where the backbone is the same (Features = Graph), and \textbf{Best Combination}, defined as the highest mean accuracy entries where backbone is different (Features $\neq$ Graph).
 \textbf{Relative Gain} corresponds to the percentage improvement of Best Combination over Best Single for the same manifold learning setting.}
\label{tab:all_datasets_acc_gain_sgc}
\renewcommand{\arraystretch}{1.15}
\setlength{\tabcolsep}{5pt}

\resizebox{\textwidth}{!}{%
\begin{tabular}{l | l l | l l l | c | c c}
\hline
\multicolumn{9}{c}{\textbf{Flowers Dataset}} \\
\hline
\textbf{Manifold} & \multicolumn{2}{c|}{\textbf{Best Single (Baseline)}}  &
\multicolumn{3}{c|}{\textbf{Best Combination}} & \textbf{Relative} & \multicolumn{2}{c}{\textbf{Statistical Tests}} \\
\cline{4-6}\cline{8-9}
\textbf{Learning} & \textbf{Feature} & \textbf{Accuracy (\%)} &
\textbf{Features} & \textbf{Graph} & \multicolumn{1}{l|}{\textbf{Accuracy (\%)}} &
\textbf{Gain} & \textbf{Wilcoxon $p$} & \textbf{Cohen's $d$} \\
\hline
None    & SWIN-TF & 97.06 $\pm$ 0.0331 & VIT-B16   & SWIN-TF & 97.33 $\pm$ 0.0339 & \textcolor{darkgreen}{\textbf{+0.28\%}} & $3.23 \times 10^{-5}$ & 0.5086 \\
LHRR    & SWIN-TF & 99.02 $\pm$ 0.0116 & VIT-B16   & SWIN-TF & 99.20 $\pm$ 0.0116 & \textcolor{darkgreen}{\textbf{+0.18\%}} & $1.78 \times 10^{-5}$ & 0.7685 \\
RDPAC   & SWIN-TF & 99.58 $\pm$ 0.0052 & CNN-SENet & SWIN-TF & 99.73 $\pm$ 0.0033 & \textcolor{darkgreen}{\textbf{+0.15\%}} & $9.04 \times 10^{-8}$ & 1.2059 \\
BFSTREE & SWIN-TF & 99.44 $\pm$ 0.0033 & ConvNeXt  & SWIN-TF & 99.50 $\pm$ 0.0000 & \textcolor{darkgreen}{\textbf{+0.06\%}} & $2.09 \times 10^{-5}$ & 0.4863 \\
\hline
\hline
\multicolumn{9}{c}{\textbf{Corel5k Dataset}} \\
\hline
\textbf{Manifold} & \multicolumn{2}{c|}{\textbf{Best Single (Baseline)}}  &
\multicolumn{3}{c|}{\textbf{Best Combination}} & \textbf{Relative} & \multicolumn{2}{c}{\textbf{Statistical Tests}} \\
\cline{4-6}\cline{8-9}
\textbf{Learning} & \textbf{Feature} & \textbf{Accuracy (\%)} &
\textbf{Features} & \textbf{Graph} & \multicolumn{1}{l|}{\textbf{Accuracy (\%)}} &
\textbf{Gain} & \textbf{Wilcoxon $p$} & \textbf{Cohen's $d$}  \\
\hline
None    & SWIN-TF & 95.89 $\pm$ 0.0276 & VIT-B16 & SWIN-TF & 96.39 $\pm$ 0.0409 & \textcolor{darkgreen}{\textbf{+0.52\%}} & $6.39 \times 10^{-8}$ & 1.2592 \\
LHRR    & SWIN-TF & 96.60 $\pm$ 0.0606 & VIT-B16 & SWIN-TF & 97.24 $\pm$ 0.0827 & \textcolor{darkgreen}{\textbf{+0.66\%}} & $5.78 \times 10^{-8}$ & 0.7767 \\
RDPAC   & SWIN-TF & 98.23 $\pm$ 0.0165 & VIT-B16 & SWIN-TF & 98.53 $\pm$ 0.0090 & \textcolor{darkgreen}{\textbf{+0.31\%}} & $2.13 \times 10^{-8}$ & 1.2293 \\
BFSTREE & SWIN-TF & 97.86 $\pm$ 0.0542 & VIT-B16 & SWIN-TF & 98.38 $\pm$ 0.0141 & \textcolor{darkgreen}{\textbf{+0.53\%}} & $2.49 \times 10^{-9}$ & 1.7120 \\
\hline
\hline
\multicolumn{9}{c}{\textbf{CUB200 Dataset}} \\
\hline
\textbf{Manifold} & \multicolumn{2}{c|}{\textbf{Best Single (Baseline)}}  &
\multicolumn{3}{c|}{\textbf{Best Combination}} & \textbf{Relative} & \multicolumn{2}{c}{\textbf{Statistical Tests}} \\
\cline{4-6}\cline{8-9}
\textbf{Learning} & \textbf{Feature} & \textbf{Accuracy (\%)} &
\textbf{Features} & \textbf{Graph} & \multicolumn{1}{l|}{\textbf{Accuracy (\%)}} &
\textbf{Gain} & \textbf{Wilcoxon $p$} &  \textbf{Cohen's $d$} \\
\hline
None    & SWIN-TF & 77.51 $\pm$ 0.0181 & VIT-B16    & SWIN-TF & 77.85 $\pm$ 0.0576 & \textcolor{darkgreen}{\textbf{+0.44\%}} & $3.36 \times 10^{-4}$ & 0.6513 \\
LHRR    & SWIN-TF & 78.92 $\pm$ 0.0054 & VIT-B16    & SWIN-TF & 80.53 $\pm$ 0.1050 & \textcolor{darkgreen}{\textbf{+2.04\%}} & $7.53 \times 10^{-10}$ & 2.7154 \\
RDPAC   & SWIN-TF & 81.86 $\pm$ 0.0232 & CNN-ResNet & SWIN-TF & 82.25 $\pm$ 0.0227 & \textcolor{darkgreen}{\textbf{+0.48\%}} & $3.14 \times 10^{-9}$ & 1.1440 \\
BFSTREE & SWIN-TF & 79.75 $\pm$ 0.0168 & VIT-B16    & SWIN-TF & 80.65 $\pm$ 0.0409 & \textcolor{darkgreen}{\textbf{+1.13\%}} & $1.76 \times 10^{-9}$ & 1.8928 \\
\hline
\end{tabular}%
}
\end{table*}

\begin{table*}[ht!]
\centering
\caption{Accuracy (\%) of single-feature, cross-combination experiments on Flowers~\cite{Flowers} dataset for \textbf{different GNN models}. The highest accuracy is highlighted in blue.}
\label{tab:flowers-networks}
\resizebox{.88\textwidth}{!}{
\begin{tabular}{|c|c|c|c|c|c|c|}
\hline
 Network & Features   & Graph  & No Manifold Learning & LHRR           & RDPAC          & BFSTREE        \\
 \hline
 \multirow{7}{*}{GCN} & CNN-ResNet & \multirow{7}{*}{SWIN-TF} & 95.38 $\pm$ 0.0248 & 98.54 $\pm$ 0.0431 & 99.71 $\pm$ 0.0052 & 99.44 $\pm$ 0.0307 \\
 & CNN-DPNet  &  & 93.99 $\pm$ 0.1042 & 98.03 $\pm$ 0.0661 & 99.69 $\pm$ 0.0122 & 99.42 $\pm$ 0.0211 \\
 & CNN-SENet  &  & 95.06 $\pm$ 0.0870 & 98.36 $\pm$ 0.0884 & \textcolor{blue}{\textbf{99.73 $\pm$ 0.0122}} & 99.34 $\pm$ 0.0270 \\
 & T2T-VIT24  &  & 94.76 $\pm$ 0.0521 & 98.30 $\pm$ 0.0462 & 99.68 $\pm$ 0.0111 & 99.38 $\pm$ 0.0692 \\
 & VIT-B16    &  & 97.29 $\pm$ 0.1101 & \textbf{99.22 $\pm$ 0.0197} & 99.49 $\pm$ 0.0040 & \textbf{99.49 $\pm$ 0.0229} \\
 & SWIN-TF    &  & \textbf{97.09 $\pm$ 0.0871} & 99.02 $\pm$ 0.0168 & 99.58 $\pm$ 0.0095 & 99.44 $\pm$ 0.0095 \\
 & ConvNeXt   &  & 94.52 $\pm$ 0.0722 & 98.31 $\pm$ 0.0583 & 99.65 $\pm$ 0.0083 & 99.43 $\pm$ 0.0391 \\
 \hline
 \multirow{7}{*}{GAT} & CNN-ResNet & \multirow{7}{*}{SWIN-TF}& 96.04 $\pm$ 0.1724 & 98.56 $\pm$ 0.0245 & \textbf{99.71 $\pm$ 0.0344} & \textbf{99.50 $\pm$ 0.0367 } \\
 & CNN-DPNet  &  & 95.17 $\pm$ 0.1316 & 98.08 $\pm$ 0.1128 & 99.69 $\pm$ 0.0228 & 99.48 $\pm$ 0.0255 \\
 & CNN-SENet  &  & 96.11 $\pm$ 0.0968 & 98.54 $\pm$ 0.0458 & 99.68 $\pm$ 0.0197 & 99.47 $\pm$ 0.0626 \\
 & T2T-VIT24  &  & 95.94 $\pm$ 0.0613 & 98.27 $\pm$ 0.0735 & 99.59 $\pm$ 0.0378 & 99.34 $\pm$ 0.0755 \\
 & VIT-B16    &  & 97.53 $\pm$ 0.1306 & \textbf{99.08 $\pm$ 0.2245} & 99.48 $\pm$ 0.0120 & 99.47 $\pm$ 0.1336 \\
 & SWIN-TF    &  & \textbf{98.07 $\pm$ 0.0470 }& 99.04 $\pm$ 0.0280 & 99.50 $\pm$ 0.0273 & 99.43 $\pm$ 0.0779 \\
 & ConvNeXt   &  & 95.73 $\pm$ 0.1811 & 98.33 $\pm$ 0.1009 & 99.61 $\pm$ 0.0246 & 99.41 $\pm$ 0.0597 \\
 \hline
 \multirow{7}{*}{APPNP} & CNN-ResNet & \multirow{7}{*}{SWIN-TF} & 95.84 $\pm$ 0.2240 & 99.11 $\pm$ 0.0548 & 99.68 $\pm$ 0.0183 & 99.57 $\pm$ 0.0228 \\
 & CNN-DPNet  &  & 95.51 $\pm$ 0.1488 & 98.71 $\pm$ 0.0484 & \textbf{99.70 $\pm$ 0.0108 } & 99.53 $\pm$ 0.0368 \\
 & CNN-SENet  &  & 96.02 $\pm$ 0.1292 & 98.97 $\pm$ 0.0329 & 99.63 $\pm$ 0.0171 & 99.57 $\pm$ 0.0189 \\
 & T2T-VIT24  &  & 95.98 $\pm$ 0.1242 & 99.01 $\pm$ 0.0286 & 99.64 $\pm$ 0.0095 & 99.53 $\pm$ 0.0235 \\
 & VIT-B16    &  & 96.10 $\pm$ 0.4908 & \textbf{99.25 $\pm$ 0.0111} & 99.42 $\pm$ 0.0266 & 99.54 $\pm$ 0.0095 \\
 & SWIN-TF    &  & \textbf{97.46 $\pm$ 0.0356} & 99.08 $\pm$ 0.0160 & 99.60 $\pm$ 0.0120 & 99.55 $\pm$ 0.0083 \\
 & ConvNeXt   &  & 95.60 $\pm$ 0.1634 & 99.06 $\pm$ 0.0189 & 99.57 $\pm$ 0.0189 & \textbf{99.57 $\pm$ 0.0065} \\
 \hline
 \multirow{7}{*}{ARMA} & CNN-ResNet & \multirow{7}{*}{SWIN-TF} & 95.38 $\pm$ 0.1652 & 98.67 $\pm$ 0.0642 & \textbf{99.70 $\pm$ 0.0158}  & 99.50 $\pm$ 0.0250   \\
 & CNN-DPNet  &  & 94.48 $\pm$ 0.1094 & 98.17 $\pm$ 0.0303 & 99.69 $\pm$ 0.0217 & 99.42 $\pm$ 0.0782 \\
 & CNN-SENet  &  & 95.44 $\pm$ 0.1526 & 98.62 $\pm$ 0.0437 & 99.66 $\pm$ 0.0300 & 99.42 $\pm$ 0.0356 \\
 & T2T-VIT24  &  & 95.31 $\pm$ 0.1206 & 98.62 $\pm$ 0.1191 & 99.67 $\pm$ 0.0250 & 99.36 $\pm$ 0.0418 \\
 & VIT-B16    &  & 97.18 $\pm$ 0.1126 & 99.07 $\pm$ 0.0565 & 99.48 $\pm$ 0.0203 & 99.49 $\pm$ 0.0261 \\
 & SWIN-TF    &  & \textbf{98.06 $\pm$ 0.0711} & \textbf{99.10 $\pm$ 0.0409} & 99.63 $\pm$ 0.0122 & \textbf{99.50 $\pm$ 0.0065}  \\
 & ConvNeXt   &  & 94.91 $\pm$ 0.0911 & 98.40 $\pm$ 0.0851 & 99.63 $\pm$ 0.0137 & 99.36 $\pm$ 0.0915 \\
\hline
\end{tabular}
}
\end{table*}

\begin{table*}[ht!]
\centering
\caption{Accuracy (\%) of single-feature, cross-combination experiments on  Corel5k~\cite{corel5k}  dataset for \textbf{different GNN models}. The highest accuracy is highlighted in blue.}
\label{tab:corel5k-networks}
\resizebox{.88\textwidth}{!}{
\begin{tabular}{|c|c|c|c|c|c|c|}
\hline
 Network & Features   & Graph   & No Manifold Learning & LHRR           & RDPAC       & BFSTREE        \\
 \hline
 \multirow{8}{*}{GCN} & CNN-ResNet & \multirow{8}{*}{SWIN-TF} 
 & 96.47 $\pm$ 0.1550 & 97.32 $\pm$ 0.1540  & 97.88 $\pm$ 0.1486 & 97.48 $\pm$ 0.1637 \\
 & CNN-DPNet  &  & 95.84 $\pm$ 0.0906 &96.89 $\pm$ 0.1504 & 97.46 $\pm$ 0.1555 & 97.13 $\pm$ 0.1758 \\
 & CNN-SENet  &  & 96.70 $\pm$ 0.0609 &97.46 $\pm$ 0.1232 & 97.82 $\pm$ 0.1148 & 97.40 $\pm$ 0.1608 \\
 & T2T-VIT24  &  & 97.03 $\pm$ 0.0809 &\textbf{97.56 $\pm$ 0.1824} & 98.00 $\pm$ 0.1137 & 97.77 $\pm$ 0.0740 \\
 & VIT-B16    &  & 96.80 $\pm$ 0.1922 &97.52 $\pm$ 0.0918 & 97.98 $\pm$ 0.1390 & \textbf{97.90 $\pm$ 0.2328} \\
 & SWIN-TF    &  & \textbf{97.72 $\pm$ 0.1403} &97.48 $\pm$ 0.1364 & \textbf{98.04 $\pm$ 0.1205} & 97.67 $\pm$ 0.1509 \\
 & DINOv2     &  & 96.86 $\pm$ 0.0917 &97.22 $\pm$ 0.2113 & 97.59 $\pm$ 0.1222 & 97.15 $\pm$ 0.2018 \\
 & ConvNeXt   &  & 97.29 $\pm$ 0.0408 &97.25 $\pm$ 0.0675 & 97.60 $\pm$ 0.1212 & 97.44 $\pm$ 0.1055 \\
 \hline
 \multirow{8}{*}{GAT} & CNN-ResNet & \multirow{8}{*}{SWIN-TF} 
 &96.52 $\pm$ 0.1319 &97.30 $\pm$ 0.1370   & 97.75 $\pm$ 0.1546 & 97.52 $\pm$ 0.0775 \\
 & CNN-DPNet  &  & 96.00 $\pm$ 0.0781 &96.92 $\pm$ 0.1183 & 97.45 $\pm$ 0.1623 & 97.34 $\pm$ 0.0948 \\
 & CNN-SENet  &  & 96.68 $\pm$ 0.0512 &97.32 $\pm$ 0.2041 & 97.67 $\pm$ 0.0874 & 97.17 $\pm$ 0.0683 \\
 & T2T-VIT24  &  & 97.04 $\pm$ 0.0581 &97.53 $\pm$ 0.0543 & 97.95 $\pm$ 0.0441 & 97.79 $\pm$ 0.0093 \\
 & VIT-B16    &  & 97.26 $\pm$ 0.0771 &\textbf{97.83 $\pm$ 0.0759} & \textbf{98.24 $\pm$ 0.0950} & \textbf{98.14 $\pm$ 0.0233} \\
 & SWIN-TF    &  & \textbf{97.66 $\pm$ 0.0743} &97.73 $\pm$ 0.0576 & 98.04 $\pm$ 0.0558 & 97.71 $\pm$ 0.0628 \\
 & DINOv2     &  & 97.28 $\pm$ 0.0621 &97.70 $\pm$ 0.1532 & 98.03 $\pm$ 0.0450 & 97.74 $\pm$ 0.0323 \\
 & ConvNeXt   &  & 97.31 $\pm$ 0.0414 &97.02 $\pm$ 0.0900 & 97.56 $\pm$ 0.0421 & 97.45 $\pm$ 0.0564 \\
 \hline
 \multirow{8}{*}{APPNP} & CNN-ResNet & \multirow{8}{*}{SWIN-TF} 
 &97.10 $\pm$ 0.0712 &97.50 $\pm$ 0.1337  & 98.22 $\pm$ 0.0683 & 97.87 $\pm$ 0.0886 \\
 & CNN-DPNet  &  &96.84 $\pm$ 0.0877 & 97.32 $\pm$ 0.0748 & 97.80 $\pm$ 0.1679 & 97.72 $\pm$ 0.0353 \\
 & CNN-SENet  &  &97.28 $\pm$ 0.0369 & 97.71 $\pm$ 0.1040 & 98.26 $\pm$ 0.0740 & 97.93 $\pm$ 0.0374 \\
 & T2T-VIT24  &  &97.61 $\pm$ 0.0714 & 97.97 $\pm$ 0.0939 & 98.34 $\pm$ 0.0169 & 98.23 $\pm$ 0.0382 \\
 & VIT-B16    &  &96.94 $\pm$ 0.1224 & 97.78 $\pm$ 0.1220 & 98.16 $\pm$ 0.0899 & 98.06 $\pm$ 0.1015 \\
 & SWIN-TF    &  &\textbf{98.19 $\pm$ 0.1122} & 98.11 $\pm$ 0.0546 & \textcolor{blue}{\textbf{98.43 $\pm$ 0.0485}} & \textbf{98.30 $\pm$ 0.0156} \\
 & DINOv2     &  &97.87 $\pm$ 0.0213 & \textbf{98.19 $\pm$ 0.0368} & 98.30 $\pm$ 0.0740 & 98.13 $\pm$ 0.0174 \\
 & ConvNeXt   &  &97.60 $\pm$ 0.0648 & 97.26 $\pm$ 0.0376 & 97.83 $\pm$ 0.0765 & 97.75 $\pm$ 0.1476 \\
 \hline
 \multirow{8}{*}{ARMA} & CNN-ResNet & \multirow{8}{*}{SWIN-TF} 
 &96.61 $\pm$ 0.0787 &97.41 $\pm$ 0.1333 & 97.74 $\pm$ 0.0412 & 97.67 $\pm$ 0.1164 \\
 & CNN-DPNet  &  &96.25 $\pm$ 0.1116 & 97.16 $\pm$ 0.1212 & 97.82 $\pm$ 0.0767 & 97.46 $\pm$ 0.1140  \\
 & CNN-SENet  &  &96.91 $\pm$ 0.0889 & 97.67 $\pm$ 0.1123 & 97.97 $\pm$ 0.1118 & 97.78 $\pm$ 0.0988 \\
 & T2T-VIT24  &  &97.13 $\pm$ 0.0649 & 97.67 $\pm$ 0.0535 & 98.09 $\pm$ 0.0803 & 98.01 $\pm$ 0.1423 \\
 & VIT-B16    &  &96.21 $\pm$ 0.1380 & 97.25 $\pm$ 0.1322 & 97.55 $\pm$ 0.2195 & 97.63 $\pm$ 0.0643 \\
 & SWIN-TF    &  &\textbf{97.88 $\pm$ 0.0653} & 97.91 $\pm$ 0.0833 & \textbf{98.23 $\pm$ 0.0591} & \textbf{98.13 $\pm$ 0.0589} \\
 & DINOv2     &  &97.56 $\pm$ 0.0618 & \textbf{97.98 $\pm$ 0.0688} & 98.17 $\pm$ 0.0444 & 98.09 $\pm$ 0.0204 \\
 & ConvNeXt   &  &97.18 $\pm$ 0.0587 & 97.32 $\pm$ 0.1198 & 97.71 $\pm$ 0.1112 & 97.57 $\pm$ 0.1076 \\
\hline
\end{tabular}
}
\end{table*}

\begin{table*}[ht!]
\centering
\caption{Accuracy (\%) of single-feature, cross-combination experiments on  CUB200~\cite{cub200} dataset for \textbf{different GNN models}. The highest accuracy is highlighted in blue.}
\label{tab:cub200-networks}
\resizebox{.88\textwidth}{!}{
\begin{tabular}{|c|c|c|c|c|c|c|}
\hline
 Network & Features   & Graph   & No Manifold Learning & LHRR    & RDPAC     & BFSTREE  \\
 \hline
  \multirow{7}{*}{GCN} & CNN-ResNet & \multirow{7}{*}{SWIN-TF} 
  & 17.64 $\pm$ 5.0767 & 38.58 $\pm$ 6.7136 & 72.09 $\pm$ 3.0904 & 65.76 $\pm$ 2.5238 \\
 & CNN-DPNet  & & 03.18 $\pm$ 0.9560 & 05.45 $\pm$ 1.2315 & 17.12 $\pm$ 6.0066 & 15.89 $\pm$ 3.5367 \\
 & CNN-SENet  & & 64.83 $\pm$ 1.7785 & 75.26 $\pm$ 0.1836 & 81.03 $\pm$ 0.1074 & 78.70 $\pm$ 0.0821  \\
 & VIT-B16    & & 33.04 $\pm$ 1.7061 & 58.45 $\pm$ 1.6373 & 77.12 $\pm$ 0.5726 & 72.72 $\pm$ 0.9333 \\
 & SWIN-TF    & & \textbf{76.47 $\pm$ 0.0931} & \textbf{78.51 $\pm$ 0.0491} & 81.30 $\pm$ 0.0776 & \textbf{79.35 $\pm$ 0.0645} \\
 & ConvNeXt   & & 72.03 $\pm$ 0.1136 & 76.92 $\pm$ 0.105  & \textbf{81.32 $\pm$ 0.0717} & 79.24 $\pm$ 0.0778 \\
 & T2T-VIT24  & & 72.29 $\pm$ 0.1507 & 76.64 $\pm$ 0.1163 & 81.10 $\pm$ 0.0416 & 78.77 $\pm$ 0.0686 \\
 \hline
 \multirow{7}{*}{GAT} & CNN-ResNet & \multirow{7}{*}{SWIN-TF} 
  & 52.60 $\pm$ 1.8634  & 66.35 $\pm$ 1.0454 & 60.37 $\pm$ 6.3609 & 64.52 $\pm$ 5.5450  \\
 & CNN-DPNet  & & 07.53 $\pm$ 1.2215 & 09.87 $\pm$ 2.4491 & 07.56 $\pm$ 1.8992 & 09.05 $\pm$ 2.2382  \\
 & CNN-SENet  & & 65.38 $\pm$ 0.8276 & 74.00 $\pm$ 0.2773 & 77.91 $\pm$ 0.4381 & 76.40 $\pm$ 0.1994  \\
 & VIT-B16    & & 38.52 $\pm$ 1.7278 & 59.34 $\pm$ 1.5749 & 67.67 $\pm$ 2.1054 & 63.62 $\pm$ 2.1208 \\
 & SWIN-TF    & & \textbf{76.08 $\pm$ 0.1542} & \textbf{78.27 $\pm$ 0.1172} & \textbf{80.46 $\pm$ 0.2204} & \textbf{78.88 $\pm$ 0.1651} \\
 & ConvNeXt   & & 71.86 $\pm$ 0.2044 & 76.43 $\pm$ 0.1536 & 80.32 $\pm$ 0.1233 & 78.20 $\pm$ 0.0819  \\
 & T2T-VIT24  & & 71.33 $\pm$ 0.4557 & 75.99 $\pm$ 0.1054 & 79.76 $\pm$ 0.2820  & 77.84 $\pm$ 0.2428 \\
 \hline
 \multirow{7}{*}{APPNP} & CNN-ResNet & \multirow{7}{*}{SWIN-TF}
  & 02.86 $\pm$ 1.3152  & 19.16 $\pm$ 3.0704 & 69.05 $\pm$ 0.8720  & 52.49 $\pm$ 3.3811 \\
 & CNN-DPNet  & & 00.54 $\pm$ 0.0708 & 00.67 $\pm$ 0.1312 & 12.23 $\pm$ 3.4327 & 03.71 $\pm$ 1.1304  \\
 & CNN-SENet  & & 66.67 $\pm$ 0.5451 & 76.49 $\pm$ 0.2134 & 79.94 $\pm$ 0.1822 & 78.02 $\pm$ 0.1393 \\
 & VIT-B16    & & 00.75 $\pm$ 0.0827 & 02.47 $\pm$ 1.0947 & 51.63 $\pm$ 1.4268 & 29.03 $\pm$ 3.9933 \\
 & SWIN-TF    & & \textbf{76.33 $\pm$ 0.1566} & \textbf{79.23 $\pm$ 0.0918} & \textbf{81.80 $\pm$ 0.0791} & \textbf{80.16 $\pm$ 0.0624} \\
 & ConvNeXt   & & 71.66 $\pm$ 0.0606 & 77.38 $\pm$ 0.1273 & 79.85 $\pm$ 0.1657 & 78.26 $\pm$ 0.0812 \\
 & T2T-VIT24  & & 73.32 $\pm$ 0.2045 & 78.03 $\pm$ 0.1607 & 80.58 $\pm$ 0.1962 & 78.72 $\pm$ 0.0587 \\
 \hline
 \multirow{7}{*}{ARMA} & CNN-ResNet & \multirow{7}{*}{SWIN-TF} 
  & 41.77 $\pm$ 4.0646 & 61.29 $\pm$ 3.5696 & 74.47 $\pm$ 2.8047 & 69.93 $\pm$ 2.6916 \\
 & CNN-DPNet  & & 08.93 $\pm$ 1.4579 & 26.23 $\pm$ 1.0546 & 36.26 $\pm$ 6.5047 & 36.90 $\pm$ 2.8655  \\
 & CNN-SENet  & & 65.78 $\pm$ 0.5155 & 73.07 $\pm$ 1.7230 & 79.42 $\pm$ 0.1673 & 76.97 $\pm$ 0.2455 \\
 & VIT-B16    & & 23.18 $\pm$ 3.2290 & 46.30 $\pm$ 5.1995 & 48.88 $\pm$ 7.8697 & 47.60 $\pm$ 3.3173  \\
 & SWIN-TF    & & \textbf{76.05 $\pm$ 0.0796} & \textbf{79.89 $\pm$ 0.1072} & \textcolor{blue}{\textbf{82.10 $\pm$ 0.0824}} & \textbf{80.62 $\pm$ 0.1279} \\
 & ConvNeXt   & & 69.81 $\pm$ 0.1981 & 75.34 $\pm$ 0.1497 & 79.73 $\pm$ 0.0496 & 77.70 $\pm$ 0.1995  \\
 & T2T-VIT24  & & 70.96 $\pm$ 0.1314 & 76.50 $\pm$ 0.1509 & 80.32 $\pm$ 0.1990 & 78.29 $\pm$ 0.1908 \\
\hline
\end{tabular}
}
\end{table*}

\begin{table*}[!ht]
\centering
\caption{Accuracy (\%) comparison between the best cross-combinations of heterogeneous features and graph obtained from different backbones and the isolated SWIN-TF baseline (feature and graph from the same backbone, without manifold learning) across different GNN models for each dataset.}
\label{tab:best_gnn_models_with_gains}
\resizebox{.93\textwidth}{!}{%
\begin{tabular}{l | c | c c c l | c}
\hline
\multicolumn{7}{c}{\textbf{Flowers Dataset}} \\
\hline
\textbf{GNN} & \textbf{Baseline (SWIN-TF)} & \multicolumn{4}{c|}{\textbf{Best Combination}} & \textbf{Relative} \\
\cline{3-6}
\textbf{Model} & \textbf{without M.L.} & \textbf{Features} & \textbf{Graph} & \textbf{M.L.} & \textbf{Accuracy (\%)} & \textbf{Gain} \\
\hline
SGC   & $97.06 \pm 0.0331$ & CNN-SENet  & SWIN-TF & RDPAC & \textbf{99.73 $\pm$ 0.0033} & \textcolor{darkgreen}{\textbf{+2.75\%}} \\
GCN   & $97.09 \pm 0.0871$ & CNN-SENet  & SWIN-TF & RDPAC & \textbf{99.73 $\pm$ 0.0122} & \textcolor{darkgreen}{\textbf{+2.72\%}} \\
GAT   & $98.07 \pm 0.0470$ & CNN-ResNet & SWIN-TF & RDPAC & $99.71 \pm 0.0344$ & \textcolor{darkgreen}{\textbf{+1.67\%}} \\
APPNP & $97.46 \pm 0.0356$ & CNN-DPNet  & SWIN-TF & RDPAC & $99.70 \pm 0.0108$ & \textcolor{darkgreen}{\textbf{+2.30\%}} \\
ARMA  & $98.06 \pm 0.0711$ & CNN-ResNet & SWIN-TF & RDPAC & $99.70 \pm 0.0158$ & \textcolor{darkgreen}{\textbf{+1.67\%}} \\
\hline
\hline
\multicolumn{7}{c}{\textbf{Corel5k Dataset}} \\
\hline
\textbf{GNN} & \textbf{Baseline (SWIN-TF)} & \multicolumn{4}{c|}{\textbf{Best Combination}} & \textbf{Relative} \\
\cline{3-6}
\textbf{Model} & \textbf{without M.L.} & \textbf{Features} & \textbf{Graph} & \textbf{M.L.} & \textbf{Accuracy (\%)} & \textbf{Gain} \\
\hline
SGC   & $95.89 \pm 0.0276$ & VIT-B16  & SWIN-TF & RDPAC & \textbf{98.53 $\pm$ 0.0090} & \textcolor{darkgreen}{\textbf{+2.75\%}} \\
GCN   & $97.72 \pm 0.1403$ & SWIN-TF  & SWIN-TF & RDPAC & $98.04 \pm 0.1205$ & \textcolor{darkgreen}{\textbf{+0.33\%}} \\
GAT   & $97.66 \pm 0.0743$ & VIT-B16  & SWIN-TF & RDPAC & $98.24 \pm 0.0950$ & \textcolor{darkgreen}{\textbf{+0.59\%}} \\
APPNP & $98.19 \pm 0.1122$ & SWIN-TF  & SWIN-TF & RDPAC & $98.43 \pm 0.0485$ & \textcolor{darkgreen}{\textbf{+0.24\%}} \\
ARMA  & $97.88 \pm 0.0653$ & SWIN-TF  & SWIN-TF & RDPAC & $98.23 \pm 0.0591$ & \textcolor{darkgreen}{\textbf{+0.36\%}} \\
\hline
\hline
\multicolumn{7}{c}{\textbf{CUB200 Dataset}} \\
\hline
\textbf{GNN} & \textbf{Baseline (SWIN-TF)} & \multicolumn{4}{c|}{\textbf{Best Combination}} & \textbf{Relative} \\
\cline{3-6}
\textbf{Model} & \textbf{without M.L.} & \textbf{Features} & \textbf{Graph} & \textbf{M.L.} & \textbf{Accuracy (\%)} & \textbf{Gain} \\
\hline
SGC   & $77.51 \pm 0.0181$ & CNN-ResNet & SWIN-TF & RDPAC & \textbf{82.25 $\pm$ 0.0227} & \textcolor{darkgreen}{\textbf{+6.12\%}} \\
GCN   & $76.47 \pm 0.0931$ & ConvNeXt   & SWIN-TF & RDPAC & $81.32 \pm 0.0717$ & \textcolor{darkgreen}{\textbf{+6.34\%}} \\
GAT   & $76.08 \pm 0.1542$ & SWIN-TF    & SWIN-TF & RDPAC & $80.46 \pm 0.2204$ & \textcolor{darkgreen}{\textbf{+5.76\%}} \\
APPNP & $76.33 \pm 0.1566$ & SWIN-TF    & SWIN-TF & RDPAC & $81.80 \pm 0.0791$ & \textcolor{darkgreen}{\textbf{+7.17\%}} \\
ARMA  & $76.05 \pm 0.0796$ & SWIN-TF    & SWIN-TF & RDPAC & $82.10 \pm 0.0824$ & \textcolor{darkgreen}{\textbf{+7.96\%}} \\
\hline
\end{tabular}%
}
\end{table*}

\begin{table*}[ht!]
\centering
\caption{Accuracy (\%) comparison of the best descriptor combinations obtained under the same experimental settings and hyperparameters for our approach versus the ManifoldGCN~\cite{valem2023graph} baseline.}
\label{tab:summary_manifold_vs_combination_all}
\resizebox{0.6\textwidth}{!}{
\begin{tabular}{|c|lc|c|}
\hline
\textbf{Dataset} & \multicolumn{2}{c|}{\textbf{ManifoldGCN~\cite{valem2023graph}}} & \textbf{Ours (Best)} \\
\hline
\multirow{4}{*}{\textbf{Flowers}}
& CNN-ResNet   & 84.40 $\pm$ 0.0572 & \multirow{4}{*}{\centering\textbf{99.73 $\pm$ 0.0033}} \\
& CNN-SENet    & 75.13 $\pm$ 0.0329 & \\
& VIT-B16  & 96.95 $\pm$ 0.0228 & \\
& SWIN-TF  & 99.58 $\pm$ 0.0052 & \\
\hline
\multirow{5}{*}{\textbf{Corel5k}}
& CNN-ResNet  & 91.95 $\pm$ 0.0212 & \multirow{5}{*}{\centering\textbf{98.53 $\pm$ 0.0090}} \\
& CNN-SENet   & 91.41 $\pm$ 0.0137 & \\
& VIT-B16     & 95.29 $\pm$ 0.0372 & \\
& DINOv2      & 95.11 $\pm$ 0.0971 & \\
& SWIN-TF     & 98.23 $\pm$ 0.0165 & \\
\hline
\multirow{4}{*}{\textbf{CUB200}}
& CNN-ResNet   & 51.81 $\pm$ 0.0340 & \multirow{4}{*}{\centering\textbf{82.25 $\pm$ 0.0227}} \\
& CNN-SENet    & 37.54 $\pm$ 0.0236 & \\
& VIT-B16  & 78.21 $\pm$ 0.0510 & \\
& SWIN-TF  & 81.86 $\pm$ 0.0232 & \\
\hline
\end{tabular}
}
\end{table*}

\begin{table*}[ht!]
\caption{Accuracy (\%) obtained with multi-feature combinations using reciprocal kNN graphs and rank aggregation on the Flowers~\cite{Flowers} dataset with SGC model. The highest accuracy is highlighted in blue.}
\label{tab:rankagg_flowers}
\centering
\resizebox{\textwidth}{!}{
\begin{tabular}{|c|c|c|c|c|c|}
\hline
 Features                   & Graph                      &  No Manifold Learning   & LHRR           & RDPAC          & BFSTREE        \\
 \hline
 \multirow{7}{*}{SWIN-TF}                    & SWIN-TF                    & \textbf{99.76 $\pm$ 0.0204} & \textbf{99.14 $\pm$ 0.0065} & \textbf{99.82 $\pm$ 0.0083} & \textbf{99.49 $\pm$ 0.0120}  \\
                     & CNN-ResNet                 & 89.26 $\pm$ 0.0472 & 87.55 $\pm$ 0.0290 & 86.53 $\pm$ 0.0692 & 85.35 $\pm$ 0.0167 \\
                     & VIT-B16                    & 97.88 $\pm$ 0.0111 & 96.02 $\pm$ 0.0065 & 97.23 $\pm$ 0.0073 & 96.79 $\pm$ 0.0116 \\
                     & SWIN-TF+VIT-B16            & 99.46 $\pm$ 0.0217 & 98.18 $\pm$ 0.0061 & 99.28 $\pm$ 0.0061 & 98.54 $\pm$ 0.0061 \\
                     & SWIN-TF+CNN-ResNet         & 99.30 $\pm$ 0.0061 & 98.67 $\pm$ 0.0217 & 99.41 $\pm$ 0.0061 & 98.22 $\pm$ 0.0137 \\
                     & VIT-B16+CNN-ResNet         & 92.86 $\pm$ 0.0211 & 86.50 $\pm$ 0.0179 & 89.29 $\pm$ 0.0235 & 87.38 $\pm$ 0.0196 \\
                     & SWIN-TF+VIT-B16+CNN-ResNet & 99.32 $\pm$ 0.0137 & 98.85 $\pm$ 0.0095 & 99.62 $\pm$ 0.0040 & 98.25 $\pm$ 0.0061 \\
 \hline
 \multirow{7}{*}{CNN-ResNet}                 & SWIN-TF                    & \textbf{97.87 $\pm$ 0.0260} & \textbf{98.94 $\pm$ 0.0131} & \textbf{99.59 $\pm$ 0.0000} & \textbf{99.52 $\pm$ 0.0120}  \\
                  & CNN-ResNet                 & 83.63 $\pm$ 0.0772 & 84.89 $\pm$ 0.0588 & 84.59 $\pm$ 0.0528 & 83.22 $\pm$ 0.0461 \\
                  & VIT-B16                    & 96.61 $\pm$ 0.0131 & 95.78 $\pm$ 0.0083 & 97.31 $\pm$ 0.0116 & 96.78 $\pm$ 0.0337 \\
                  & SWIN-TF+VIT-B16            & 97.02 $\pm$ 0.0438 & 96.34 $\pm$ 0.043  & 98.87 $\pm$ 0.0065 & 98.54 $\pm$ 0.0137 \\
                  & SWIN-TF+CNN-ResNet         & 96.87 $\pm$ 0.0266 & 97.07 $\pm$ 0.0486 & 99.16 $\pm$ 0.0120 & 98.08 $\pm$ 0.0168 \\
                  & VIT-B16+CNN-ResNet         & 85.82 $\pm$ 0.0887 & 84.05 $\pm$ 0.0347 & 87.21 $\pm$ 0.0951 & 85.61 $\pm$ 0.0769 \\
                  & SWIN-TF+VIT-B16+CNN-ResNet & 97.06 $\pm$ 0.0381 & 97.19 $\pm$ 0.0176 & 99.47 $\pm$ 0.0095 & 98.18 $\pm$ 0.0122 \\
 \hline
 \multirow{7}{*}{VIT-B16}                    & SWIN-TF                    & \textbf{99.13 $\pm$ 0.0197} & \textbf{99.30 $\pm$ 0.0040} & \textbf{99.73 $\pm$ 0.0040} & \textbf{99.55 $\pm$ 0.0040}  \\
                     & CNN-ResNet                 & 88.99 $\pm$ 0.0347 & 88.39 $\pm$ 0.0276 & 87.53 $\pm$ 0.0216 & 86.08 $\pm$ 0.0520  \\
                     & VIT-B16                    & 96.58 $\pm$ 0.0131 & 95.84 $\pm$ 0.0228 & 97.21 $\pm$ 0.0189 & 96.78 $\pm$ 0.0317 \\
                     & SWIN-TF+VIT-B16            & 98.76 $\pm$ 0.0245 & 98.37 $\pm$ 0.0273 & 99.24 $\pm$ 0.0040 & 98.64 $\pm$ 0.0061 \\
                     & SWIN-TF+CNN-ResNet         & 98.81 $\pm$ 0.0158 & 98.60 $\pm$ 0.0167 & 99.30 $\pm$ 0.0061 & 98.57 $\pm$ 0.0089 \\
                     & VIT-B16+CNN-ResNet         & 92.15 $\pm$ 0.0546 & 87.71 $\pm$ 0.0480 & 90.24 $\pm$ 0.0108 & 87.88 $\pm$ 0.0340  \\
                     & SWIN-TF+VIT-B16+CNN-ResNet & 98.89 $\pm$ 0.0089 & 98.67 $\pm$ 0.0133 & 99.48 $\pm$ 0.0061 & 98.52 $\pm$ 0.0052 \\
 \hline
 \multirow{7}{*}{SWIN-TF+VIT-B16}            & SWIN-TF                    & \textbf{99.54 $\pm$ 0.0280} & \textbf{99.26 $\pm$ 0.0089} & \textbf{99.81 $\pm$ 0.0073} & \textbf{99.54 $\pm$ 0.0061} \\
             & CNN-ResNet                 & 88.78 $\pm$ 0.0275 & 88.02 $\pm$ 0.0329 & 87.36 $\pm$ 0.0299 & 85.85 $\pm$ 0.0440  \\
             & VIT-B16                    & 97.69 $\pm$ 0.0381 & 96.14 $\pm$ 0.0083 & 97.49 $\pm$ 0.0186 & 97.06 $\pm$ 0.0167 \\
             & SWIN-TF+VIT-B16            & 99.18 $\pm$ 0.0235 & 98.15 $\pm$ 0.0323 & 99.25 $\pm$ 0.0080 & 98.60 $\pm$ 0.0033  \\
             & SWIN-TF+CNN-ResNet         & 99.01 $\pm$ 0.0266 & 98.63 $\pm$ 0.0152 & 99.42 $\pm$ 0.0120 & 98.49 $\pm$ 0.0098 \\
             & VIT-B16+CNN-ResNet         & 91.79 $\pm$ 0.0250 & 87.55 $\pm$ 0.0168 & 89.73 $\pm$ 0.0168 & 88.06 $\pm$ 0.0209 \\
             & SWIN-TF+VIT-B16+CNN-ResNet & 99.02 $\pm$ 0.0108 & 98.79 $\pm$ 0.0174 & 99.61 $\pm$ 0.0065 & 98.57 $\pm$ 0.0182 \\
 \hline
 \multirow{7}{*}{SWIN-TF+CNN-ResNet}         & SWIN-TF                    & \textbf{99.43 $\pm$ 0.0337} & \textbf{99.23 $\pm$ 0.0095} & \textbf{99.84 $\pm$ 0.0033} & \textbf{99.51 $\pm$ 0.0083} \\
          & CNN-ResNet                 & 87.29 $\pm$ 0.0603 & 87.24 $\pm$ 0.0385 & 86.49 $\pm$ 0.0908 & 85.12 $\pm$ 0.0385 \\
          & VIT-B16                    & 97.80 $\pm$ 0.0111 & 96.17 $\pm$ 0.0061 & 97.68 $\pm$ 0.0095 & 97.03 $\pm$ 0.0108 \\
          & SWIN-TF+VIT-B16            & 98.96 $\pm$ 0.0246 & 98.12 $\pm$ 0.0179 & 99.22 $\pm$ 0.0061 & 98.62 $\pm$ 0.0122 \\
          & SWIN-TF+CNN-ResNet         & 98.66 $\pm$ 0.0261 & 98.53 $\pm$ 0.0040 & 99.33 $\pm$ 0.0133 & 98.40 $\pm$ 0.0171  \\
          & VIT-B16+CNN-ResNet         & 89.49 $\pm$ 0.0645 & 86.38 $\pm$ 0.0514 & 89.00 $\pm$ 0.0248 & 87.54 $\pm$ 0.0497 \\
          & SWIN-TF+VIT-B16+CNN-ResNet & 98.70 $\pm$ 0.0430 & 98.70 $\pm$ 0.0158 & 99.55 $\pm$ 0.0176 & 98.47 $\pm$ 0.0095 \\
 \hline
 \multirow{7}{*}{VIT-B16+CNN-ResNet}         & SWIN-TF                    & \textbf{99.10 $\pm$ 0.0211} & \textbf{99.19 $\pm$ 0.0122} & \textbf{99.81 $\pm$ 0.0083} & \textbf{99.53 $\pm$ 0.0083} \\
  & CNN-ResNet                 & 84.94 $\pm$ 0.1048 & 86.71 $\pm$ 0.0213 & 85.95 $\pm$ 0.0475 & 84.17 $\pm$ 0.0443 \\
  & VIT-B16                    & 97.31 $\pm$ 0.0508 & 96.08 $\pm$ 0.0231 & 97.60 $\pm$ 0.0191 & 96.94 $\pm$ 0.0083 \\
  & SWIN-TF+VIT-B16            & 98.46 $\pm$ 0.0340 & 98.07 $\pm$ 0.0120 & 99.28 $\pm$ 0.0095 & 98.70 $\pm$ 0.0083  \\
  & SWIN-TF+CNN-ResNet         & 98.12 $\pm$ 0.0452 & 98.01 $\pm$ 0.0288 & 99.34 $\pm$ 0.0065 & 98.58 $\pm$ 0.0152 \\
  & VIT-B16+CNN-ResNet         & 84.97 $\pm$ 0.1080 & 85.01 $\pm$ 0.0581 & 88.27 $\pm$ 0.0280 & 86.28 $\pm$ 0.0475 \\
  & SWIN-TF+VIT-B16+CNN-ResNet & 98.26 $\pm$ 0.0434 & 98.02 $\pm$ 0.0316 & 99.60 $\pm$ 0.0146 & 98.66 $\pm$ 0.0122 \\
 \hline
 \multirow{7}{*}{SWIN-TF+VIT-B16+CNN-ResNet} & SWIN-TF                    & \textbf{99.53 $\pm$ 0.0219} & \textbf{99.23 $\pm$ 0.0089} & \textcolor{blue}{\textbf{99.85 $\pm$ 0.0040}} & \textbf{99.51 $\pm$ 0.0083} \\
  & CNN-ResNet                 & 87.62 $\pm$ 0.0196 & 87.46 $\pm$ 0.1051 & 86.57 $\pm$ 0.1100 & 85.12 $\pm$ 0.0395 \\
  & VIT-B16                    & 97.64 $\pm$ 0.0176 & 96.10 $\pm$ 0.0235 & 97.37 $\pm$ 0.0213 & 96.94 $\pm$ 0.0174 \\
  & SWIN-TF+VIT-B16            & 99.14 $\pm$ 0.0308 & 98.13 $\pm$ 0.0122 & 99.26 $\pm$ 0.0073 & 98.60 $\pm$ 0.0095  \\
  & SWIN-TF+CNN-ResNet         & 98.90 $\pm$ 0.0240 & 98.56 $\pm$ 0.0158 & 99.40 $\pm$ 0.0203 & 98.41 $\pm$ 0.0255 \\
  & VIT-B16+CNN-ResNet         & 90.08 $\pm$ 0.0317 & 86.55 $\pm$ 0.0528 & 89.19 $\pm$ 0.0381 & 87.65 $\pm$ 0.0509 \\
  & SWIN-TF+VIT-B16+CNN-ResNet & 98.95 $\pm$ 0.0155 & 98.77 $\pm$ 0.0349 & 99.60 $\pm$ 0.0120 & 98.46 $\pm$ 0.0193 \\
\hline
\end{tabular}
}
\end{table*}

\begin{table*}[ht!]
\caption{Accuracy (\%) obtained with multi-feature combinations using reciprocal kNN graphs and rank aggregation on the Corel5k~\cite{corel5k} dataset with SGC model. The highest accuracy is highlighted in blue.}
\label{tab:rankagg_corel5k}
\centering
\resizebox{\textwidth}{!}{
\begin{tabular}{|c|c|c|c|c|c|}
\hline
 Features                   & Graph                      &  No Manifold Learning   & LHRR           & RDPAC          & BFSTREE        \\
 \hline
 \multirow{7}{*}{VIT-B16}                & VIT-B16                & 95.11 $\pm$ 0.0215 & 95.94 $\pm$ 0.0392 & 95.64 $\pm$ 0.0343 & 95.61 $\pm$ 0.0148 \\
                 & DINOv2                 & 96.82 $\pm$ 0.0108 & 96.73 $\pm$ 0.0115 & 96.77 $\pm$ 0.0168 & 96.78 $\pm$ 0.0107 \\
                 & SWIN-TF                & \textbf{97.93 $\pm$ 0.0198} & \textbf{98.19 $\pm$ 0.0102} & \textcolor{blue}{\textbf{98.52 $\pm$ 0.0103}} & \textbf{98.38 $\pm$ 0.0045} \\
                 & VIT-B16+DINOv2         & 96.64 $\pm$ 0.0134 & 96.34 $\pm$ 0.0128 & 96.45 $\pm$ 0.0195 & 96.42 $\pm$ 0.0202 \\
                 & SWIN-TF+VIT-B16        & 97.41 $\pm$ 0.0068 & 97.39 $\pm$ 0.0152 & 97.71 $\pm$ 0.0046 & 97.76 $\pm$ 0.0133 \\
                 & SWIN-TF+DINOv2         & 97.04 $\pm$ 0.0087 & 96.83 $\pm$ 0.0064 & 96.77 $\pm$ 0.0117 & 96.86 $\pm$ 0.0197 \\
                 & SWIN-TF+VIT-B16+DINOv2 & 97.01 $\pm$ 0.0084 & 96.85 $\pm$ 0.0064 & 96.86 $\pm$ 0.0155 & 96.88 $\pm$ 0.0202 \\
 \hline
 \multirow{7}{*}{DINOv2}                 & VIT-B16                & 95.57 $\pm$ 0.0485 & 95.39 $\pm$ 0.0517 & 94.63 $\pm$ 0.2286 & 94.70 $\pm$ 0.0697  \\
                  & DINOv2                 & 95.34 $\pm$ 0.0298 & 95.70 $\pm$ 0.0316 & 95.56 $\pm$ 0.0584 & 95.23 $\pm$ 0.0797 \\
                  & SWIN-TF                & \textbf{97.41 $\pm$ 0.0257} & \textbf{97.62 $\pm$ 0.0731} & \textbf{97.88 $\pm$ 0.1169} & \textbf{97.54 $\pm$ 0.0676} \\
                  & VIT-B16+DINOv2         & 94.64 $\pm$ 0.0237 & 94.99 $\pm$ 0.0779 & 94.85 $\pm$ 0.0389 & 94.76 $\pm$ 0.0821 \\
                  & SWIN-TF+VIT-B16        & 97.05 $\pm$ 0.0319 & 96.75 $\pm$ 0.0487 & 96.79 $\pm$ 0.1050 & 97.12 $\pm$ 0.0391 \\
                  & SWIN-TF+DINOv2         & 95.47 $\pm$ 0.0520 & 95.71 $\pm$ 0.0794 & 95.37 $\pm$ 0.0742 & 95.36 $\pm$ 0.0717 \\
                  & SWIN-TF+VIT-B16+DINOv2 & 95.46 $\pm$ 0.0659 & 95.77 $\pm$ 0.0895 & 95.50 $\pm$ 0.0804 & 95.36 $\pm$ 0.0549 \\
 \hline
 \multirow{7}{*}{SWIN-TF}                & VIT-B16                & 95.98 $\pm$ 0.0323 & 95.50 $\pm$ 0.0070 & 95.42 $\pm$ 0.0106 & 95.21 $\pm$ 0.0694 \\
                 & DINOv2                 & 96.34 $\pm$ 0.0491 & 96.08 $\pm$ 0.0044 & 96.34 $\pm$ 0.0073 & 96.19 $\pm$ 0.0610  \\
                 & SWIN-TF                & \textbf{98.01 $\pm$ 0.0323} & \textbf{97.81 $\pm$ 0.0762} & \textbf{98.34 $\pm$ 0.0274} & \textbf{97.94 $\pm$ 0.0567} \\
                 & VIT-B16+DINOv2         & 96.13 $\pm$ 0.1079 & 95.89 $\pm$ 0.1077 & 95.85 $\pm$ 0.0045 & 95.78 $\pm$ 0.0637 \\
                 & SWIN-TF+VIT-B16        & 97.61 $\pm$ 0.0770 & 97.29 $\pm$ 0.0117 & 97.21 $\pm$ 0.0283 & 97.57 $\pm$ 0.0146 \\
                 & SWIN-TF+DINOv2         & 96.66 $\pm$ 0.0625 & 96.44 $\pm$ 0.1031 & 96.42 $\pm$ 0.0184 & 96.07 $\pm$ 0.0595 \\
                 & SWIN-TF+VIT-B16+DINOv2 & 96.66 $\pm$ 0.0744 & 96.40 $\pm$ 0.1085 & 96.48 $\pm$ 0.0575 & 96.18 $\pm$ 0.0644 \\
 \hline
 \multirow{7}{*}{VIT-B16+DINOv2}         & VIT-B16                & 95.00 $\pm$ 0.0717 & 94.97 $\pm$ 0.1167 & 94.42 $\pm$ 0.0931 & 94.33 $\pm$ 0.1407 \\
 & DINOv2                 & 94.87 $\pm$ 0.1089 & 95.30 $\pm$ 0.1702 & 95.14 $\pm$ 0.1837 & 95.07 $\pm$ 0.1281 \\
 & SWIN-TF                & \textbf{96.48 $\pm$ 0.0942} & \textbf{97.12 $\pm$ 0.1322} & \textbf{97.36 $\pm$ 0.1463} & \textbf{97.45 $\pm$ 0.0856} \\
 & VIT-B16+DINOv2         & 93.96 $\pm$ 0.1089 & 94.39 $\pm$ 0.0842 & 94.50 $\pm$ 0.1282 & 94.51 $\pm$ 0.1056 \\
 & SWIN-TF+VIT-B16        & 96.03 $\pm$ 0.0975 & 96.07 $\pm$ 0.0595 & 96.29 $\pm$ 0.1589 & 96.65 $\pm$ 0.0683 \\
 & SWIN-TF+DINOv2         & 94.76 $\pm$ 0.0984 & 95.05 $\pm$ 0.1208 & 94.82 $\pm$ 0.0540 & 94.98 $\pm$ 0.1366 \\
 & SWIN-TF+VIT-B16+DINOv2 & 94.71 $\pm$ 0.0971 & 95.13 $\pm$ 0.1930 & 95.04 $\pm$ 0.0943 & 94.95 $\pm$ 0.1117 \\
 \hline
 \multirow{7}{*}{SWIN-TF+VIT-B16}        & VIT-B16                & 96.00 $\pm$ 0.0389 & 95.90 $\pm$ 0.1026 & 95.64 $\pm$ 0.1250 & 95.77 $\pm$ 0.0763 \\
 & DINOv2                 & 96.74 $\pm$ 0.0505 & 96.68 $\pm$ 0.0993 & 96.77 $\pm$ 0.0543 & 96.57 $\pm$ 0.0634 \\
 & SWIN-TF                & \textbf{97.97 $\pm$ 0.0432} & \textbf{98.04 $\pm$ 0.0661} & \textbf{98.37 $\pm$ 0.0798} & \textbf{98.34 $\pm$ 0.0516} \\
 & VIT-B16+DINOv2         & 96.51 $\pm$ 0.0404 & 96.16 $\pm$ 0.0651 & 96.04 $\pm$ 0.0485 & 96.00 $\pm$ 0.0582  \\
 & SWIN-TF+VIT-B16        & 97.44 $\pm$ 0.0711 & 97.38 $\pm$ 0.0160 & 97.63 $\pm$ 0.1198 & 97.69 $\pm$ 0.0393 \\
 & SWIN-TF+DINOv2         & 96.89 $\pm$ 0.0521 & 96.62 $\pm$ 0.0852 & 96.51 $\pm$ 0.0525 & 96.51 $\pm$ 0.0466 \\
 & SWIN-TF+VIT-B16+DINOv2 & 96.87 $\pm$ 0.0131 & 96.64 $\pm$ 0.0694 & 96.56 $\pm$ 0.0382 & 96.50 $\pm$ 0.0710   \\
 \hline
 \multirow{7}{*}{SWIN-TF+DINOv2}         & VIT-B16                & 95.45 $\pm$ 0.0784 & 94.98 $\pm$ 0.1163 & 94.64 $\pm$ 0.1430 & 94.61 $\pm$ 0.1504 \\
 & DINOv2                 & 95.30 $\pm$ 0.0702 & 95.51 $\pm$ 0.1464 & 95.60 $\pm$ 0.1438 & 95.28 $\pm$ 0.0737 \\
 & SWIN-TF                & \textbf{96.99 $\pm$ 0.1227} & \textbf{97.28 $\pm$ 0.0677} & \textbf{97.75 $\pm$ 0.2602} & \textbf{97.53 $\pm$ 0.1127} \\
 & VIT-B16+DINOv2         & 94.51 $\pm$ 0.1443 & 94.92 $\pm$ 0.2022 & 94.93 $\pm$ 0.0885 & 94.88 $\pm$ 0.1135 \\
 & SWIN-TF+VIT-B16        & 96.45 $\pm$ 0.0504 & 96.47 $\pm$ 0.1226 & 96.48 $\pm$ 0.0847 & 96.84 $\pm$ 0.1669 \\
 & SWIN-TF+DINOv2         & 95.38 $\pm$ 0.0642 & 95.47 $\pm$ 0.0937 & 95.48 $\pm$ 0.1255 & 95.24 $\pm$ 0.1483 \\
 & SWIN-TF+VIT-B16+DINOv2 & 95.37 $\pm$ 0.1059 & 95.54 $\pm$ 0.1089 & 95.41 $\pm$ 0.1620 & 95.23 $\pm$ 0.1955 \\
 \hline
 \multirow{7}{*}{SWIN-TF+VIT-B16+DINOv2} & VIT-B16                & 95.36 $\pm$ 0.0701 & 95.03 $\pm$ 0.1631 & 94.47 $\pm$ 0.1042 & 94.74 $\pm$ 0.1470  \\
 & DINOv2                 & 95.34 $\pm$ 0.0806 & 95.56 $\pm$ 0.1245 & 95.68 $\pm$ 0.1567 & 95.34 $\pm$ 0.1266 \\
 & SWIN-TF                & \textbf{97.06 $\pm$ 0.0812} & \textbf{97.36 $\pm$ 0.0627} & \textbf{97.74 $\pm$ 0.1474} & \textbf{97.52 $\pm$ 0.0654} \\
 & VIT-B16+DINOv2         & 94.66 $\pm$ 0.0868 & 94.83 $\pm$ 0.0490 & 94.82 $\pm$ 0.0833 & 94.75 $\pm$ 0.1345 \\
 & SWIN-TF+VIT-B16        & 96.75 $\pm$ 0.1241 & 96.53 $\pm$ 0.1229 & 96.69 $\pm$ 0.0551 & 96.98 $\pm$ 0.1579 \\
 & SWIN-TF+DINOv2         & 95.38 $\pm$ 0.0660 & 95.50 $\pm$ 0.0916 & 95.28 $\pm$ 0.1160 & 95.26 $\pm$ 0.0881 \\
 & SWIN-TF+VIT-B16+DINOv2 & 95.40 $\pm$ 0.1133 & 95.68 $\pm$ 0.1476 & 95.24 $\pm$ 0.1327 & 95.34 $\pm$ 0.1060  \\
\hline
\end{tabular}
}
\end{table*}

\begin{table*}[ht!]
\caption{Accuracy (\%) obtained with multi-feature combinations using reciprocal kNN graphs and rank aggregation on the CUB200~\cite{cub200} dataset with SGC model. The highest accuracy is highlighted in blue.}
\label{tab:rankagg_cub200}
\centering
\resizebox{\textwidth}{!}{
\begin{tabular}{|c|c|c|c|c|c|}
\hline
 Features                   & Graph                      & No Manifold Learning & LHRR           & RDPAC          & BFSTREE        \\
  \hline
 \multirow{7}{*}{CNN-ResNet}                 & CNN-ResNet                 & 53.84 $\pm$ 0.0174 & 52.13 $\pm$ 0.0376 & 52.94 $\pm$ 0.0242 & 52.71 $\pm$ 0.0366 \\
     & SWIN-TF                    & 77.95 $\pm$ 0.0085 & \textbf{79.05 $\pm$ 0.0381} & \textbf{82.65 $\pm$ 0.0181} & \textbf{80.72 $\pm$ 0.0228} \\
     & VIT-B16                    & \textbf{78.66 $\pm$ 0.0532} & 77.58 $\pm$ 0.0180 & 79.30 $\pm$ 0.0097 & 78.64 $\pm$ 0.0065 \\
     & SWIN-TF+CNN-ResNet         & 76.31 $\pm$ 0.0352 & 73.08 $\pm$ 0.0277 & 80.08 $\pm$ 0.0204 & 77.66 $\pm$ 0.0387 \\
     & SWIN-TF+VIT-B16            & 77.58 $\pm$ 0.0428 & 74.93 $\pm$ 0.0484 & 81.57 $\pm$ 0.0292 & 79.45 $\pm$ 0.0271 \\
     & VIT-B16+CNN-ResNet         & 64.72 $\pm$ 0.0426 & 60.80 $\pm$ 0.0256 & 66.02 $\pm$ 0.0436 & 63.76 $\pm$ 0.0176 \\
     & SWIN-TF+VIT-B16+CNN-ResNet & 77.28 $\pm$ 0.0332 & 73.83 $\pm$ 0.0409 & 81.00 $\pm$ 0.0163 & 78.68 $\pm$ 0.0239 \\
 \hline
 \multirow{7}{*}{SWIN-TF}                    & CNN-ResNet                 & 63.60 $\pm$ 0.0079 & 55.16 $\pm$ 0.0208 & 55.46 $\pm$ 0.0158 & 54.96 $\pm$ 0.0067 \\
  & SWIN-TF                    & \textbf{82.05 $\pm$ 0.0067} & \textbf{80.54 $\pm$ 0.0194} & \textbf{82.61 $\pm$ 0.0154} & \textbf{80.75 $\pm$ 0.0199} \\
  & VIT-B16                    & 81.02 $\pm$ 0.0141 & 78.74 $\pm$ 0.0212 & 79.55 $\pm$ 0.0152 & 78.66 $\pm$ 0.0155 \\
  & SWIN-TF+CNN-ResNet         & 80.28 $\pm$ 0.0027 & 78.40 $\pm$ 0.0285 & 80.28 $\pm$ 0.0057 & 78.39 $\pm$ 0.0089 \\
  & SWIN-TF+VIT-B16            & 81.40 $\pm$ 0.0071 & 79.63 $\pm$ 0.0231 & 81.61 $\pm$ 0.0119 & 79.59 $\pm$ 0.0075 \\
  & VIT-B16+CNN-ResNet         & 72.83 $\pm$ 0.0094 & 67.14 $\pm$ 0.0081 & 68.46 $\pm$ 0.0118 & 66.37 $\pm$ 0.0087 \\
  & SWIN-TF+VIT-B16+CNN-ResNet & 81.19 $\pm$ 0.0046 & 79.35 $\pm$ 0.0059 & 81.29 $\pm$ 0.0166 & 79.34 $\pm$ 0.0129 \\
  \hline
 \multirow{7}{*}{VIT-B16}                    & CNN-ResNet                 & 61.85 $\pm$ 0.0503 & 56.19 $\pm$ 0.0420 & 55.66 $\pm$ 0.0345 & 53.97 $\pm$ 0.0916 \\
  & SWIN-TF                    & \textbf{81.66 $\pm$ 0.0182} & \textbf{81.52 $\pm$ 0.0383} & \textcolor{blue}{\textbf{83.33 $\pm$ 0.0376}} & \textbf{81.61 $\pm$ 0.0424} \\
  & VIT-B16                    & 78.35 $\pm$ 0.0184 & 78.54 $\pm$ 0.0177 & 79.32 $\pm$ 0.0214 & 78.25 $\pm$ 0.0471 \\
  & SWIN-TF+CNN-ResNet         & 79.71 $\pm$ 0.0129 & 79.15 $\pm$ 0.0304 & 81.04 $\pm$ 0.0160 & 79.06 $\pm$ 0.0295 \\
  & SWIN-TF+VIT-B16            & 80.87 $\pm$ 0.0093 & 80.49 $\pm$ 0.0299 & 82.32 $\pm$ 0.0350 & 80.30 $\pm$ 0.0208  \\
  & VIT-B16+CNN-ResNet         & 69.84 $\pm$ 0.0240 & 67.15 $\pm$ 0.0506 & 67.76 $\pm$ 0.0277 & 65.20 $\pm$ 0.0238  \\
  & SWIN-TF+VIT-B16+CNN-ResNet & 80.52 $\pm$ 0.0163 & 79.91 $\pm$ 0.0133 & 81.86 $\pm$ 0.0159 & 80.02 $\pm$ 0.0223 \\
  \hline
 \multirow{7}{*}{SWIN-TF+CNN-ResNet}         & CNN-ResNet                 & 62.29 $\pm$ 0.0047 & 55.49 $\pm$ 0.0024 & 55.78 $\pm$ 0.0017 & 55.21 $\pm$ 0.0048 \\
  & SWIN-TF                    & \textbf{81.55 $\pm$ 0.0043} & \textbf{80.78 $\pm$ 0.0035} & \textbf{82.86 $\pm$ 0.0041} & \textbf{81.15 $\pm$ 0.0031} \\
  & VIT-B16                    & 80.84 $\pm$ 0.0017 & 79.06 $\pm$ 0.0067 & 79.86 $\pm$ 0.0019 & 79.01 $\pm$ 0.0050  \\
  & SWIN-TF+CNN-ResNet         & 79.69 $\pm$ 0.0026 & 78.16 $\pm$ 0.0027 & 80.38 $\pm$ 0.0070 & 78.57 $\pm$ 0.0036 \\
  & SWIN-TF+VIT-B16            & 80.77 $\pm$ 0.0040 & 79.52 $\pm$ 0.0097 & 81.89 $\pm$ 0.0037 & 79.74 $\pm$ 0.0020  \\
  & VIT-B16+CNN-ResNet         & 71.13 $\pm$ 0.0063 & 66.58 $\pm$ 0.0055 & 68.06 $\pm$ 0.0052 & 66.00 $\pm$ 0.0080   \\
  & SWIN-TF+VIT-B16+CNN-ResNet & 80.51 $\pm$ 0.0041 & 79.01 $\pm$ 0.0027 & 81.45 $\pm$ 0.0034 & 79.44 $\pm$ 0.0024 \\
  \hline
 \multirow{7}{*}{SWIN-TF+VIT-B16}            & CNN-ResNet                 & 63.70 $\pm$ 0.0078 & 56.12 $\pm$ 0.0045 & 56.52 $\pm$ 0.0068 & 55.68 $\pm$ 0.0043 \\
 & SWIN-TF                    & \textbf{81.79 $\pm$ 0.0064} & \textbf{80.91 $\pm$ 0.0026} & \textbf{82.96 $\pm$ 0.0030} & \textbf{81.28 $\pm$ 0.0064} \\
 & VIT-B16                    & 80.87 $\pm$ 0.0047 & 79.20 $\pm$ 0.0032 & 79.92 $\pm$ 0.0046 & 79.01 $\pm$ 0.0055 \\
 & SWIN-TF+CNN-ResNet         & 79.91 $\pm$ 0.0044 & 78.44 $\pm$ 0.0055 & 80.54 $\pm$ 0.0063 & 78.70 $\pm$ 0.0033  \\
 & SWIN-TF+VIT-B16            & 81.08 $\pm$ 0.0035 & 79.82 $\pm$ 0.0037 & 81.92 $\pm$ 0.0063 & 79.80 $\pm$ 0.0069  \\
 & VIT-B16+CNN-ResNet         & 72.52 $\pm$ 0.0105 & 67.42 $\pm$ 0.0053 & 68.87 $\pm$ 0.0097 & 66.82 $\pm$ 0.0055 \\
 & SWIN-TF+VIT-B16+CNN-ResNet & 80.73 $\pm$ 0.0061 & 79.33 $\pm$ 0.0094 & 81.60 $\pm$ 0.0042 & 79.62 $\pm$ 0.0027 \\
  \hline
 \multirow{7}{*}{VIT-B16+CNN-ResNet}         & CNN-ResNet                 & 56.41 $\pm$ 0.0176 & 53.97 $\pm$ 0.0125 & 54.00 $\pm$ 0.0117 & 53.04 $\pm$ 0.0097 \\
  & SWIN-TF                    & \textbf{79.69 $\pm$ 0.0094} & \textbf{80.97 $\pm$ 0.0085} & \textbf{82.91 $\pm$ 0.0054} & \textbf{81.37 $\pm$ 0.0099} \\
  & VIT-B16                    & 78.27 $\pm$ 0.0067 & 78.32 $\pm$ 0.0061 & 79.48 $\pm$ 0.0070 & 78.49 $\pm$ 0.0035 \\
  & SWIN-TF+CNN-ResNet         & 77.42 $\pm$ 0.0070 & 76.99 $\pm$ 0.0074 & 80.37 $\pm$ 0.0069 & 78.36 $\pm$ 0.0057 \\
  & SWIN-TF+VIT-B16            & 79.01 $\pm$ 0.0066 & 78.70 $\pm$ 0.0077 & 81.84 $\pm$ 0.0111 & 79.90 $\pm$ 0.0052  \\
  & VIT-B16+CNN-ResNet         & 64.87 $\pm$ 0.0032 & 63.69 $\pm$ 0.0051 & 65.71 $\pm$ 0.0044 & 63.62 $\pm$ 0.0262 \\
  & SWIN-TF+VIT-B16+CNN-ResNet & 78.33 $\pm$ 0.0067 & 77.84 $\pm$ 0.0109 & 81.37 $\pm$ 0.0088 & 79.29 $\pm$ 0.0084 \\
  \hline
 \multirow{7}{*}{SWIN-TF+VIT-B16+CNN-ResNet} & CNN-ResNet                 & 62.59 $\pm$ 0.0057 & 55.56 $\pm$ 0.0069 & 55.80 $\pm$ 0.0095 & 55.25 $\pm$ 0.0066 \\
  & SWIN-TF                    & \textbf{81.86 $\pm$ 0.0058} & \textbf{80.86 $\pm$ 0.0084} & \textbf{82.91 $\pm$ 0.0064} & \textbf{81.19 $\pm$ 0.0032} \\
  & VIT-B16                    & 80.83 $\pm$ 0.0082 & 79.06 $\pm$ 0.0085 & 79.83 $\pm$ 0.0079 & 78.95 $\pm$ 0.0036 \\
  & SWIN-TF+CNN-ResNet         & 79.91 $\pm$ 0.0043 & 78.32 $\pm$ 0.0038 & 80.46 $\pm$ 0.0088 & 78.61 $\pm$ 0.0057 \\
  & SWIN-TF+VIT-B16            & 81.03 $\pm$ 0.0059 & 79.74 $\pm$ 0.0054 & 81.94 $\pm$ 0.0041 & 79.77 $\pm$ 0.0089 \\
  & VIT-B16+CNN-ResNet         & 71.28 $\pm$ 0.0067 & 66.69 $\pm$ 0.0098 & 68.04 $\pm$ 0.0028 & 66.02 $\pm$ 0.0132 \\
  & SWIN-TF+VIT-B16+CNN-ResNet & 80.74 $\pm$ 0.0038 & 79.19 $\pm$ 0.0050 & 81.50 $\pm$ 0.0045 & 79.47 $\pm$ 0.0028 \\
\hline
\end{tabular}
}
\end{table*}

\begin{table*}[!ht]
\centering
\caption{Accuracy (\%) for multi-feature experiments on SGC comparing best single results without manifold learning (M.L.) and best combinations for each dataset.
Relative gain is computed as the percentage improvement of the best combination over the best single result without M.L.}
\label{tab:best_single_vs_combination_summary}
\renewcommand{\arraystretch}{1.15}
\setlength{\tabcolsep}{5pt}

\resizebox{\textwidth}{!}{%
\begin{tabular}{l | l l | l l l l | c | c c}
\hline
\textbf{Dataset} &
\multicolumn{2}{c|}{\textbf{Best Single without M.L.}} &
\multicolumn{4}{c|}{\textbf{Best Combination}} &
\textbf{Relative} &
\multicolumn{2}{c}{\textbf{Statistical Tests}} \\
\cline{4-7}\cline{9-10}
 & \textbf{Feature} & \textbf{Accuracy (\%)} &
\textbf{Features} & \textbf{Graph} & \textbf{M.L.} & \multicolumn{1}{l|}{\textbf{Accuracy (\%)}} &
\textbf{Gain} & \textbf{Wilcoxon $p$} & \textbf{Cohen's $d$} \\
\hline
Flowers &
SWIN-TF & 99.76 $\pm$ 0.0204 &
\begin{tabular}[c]{@{}l@{}}
SWIN-TF\\
+ VIT-B16\\
+ CNN-ResNet
\end{tabular}
& SWIN-TF & RDPAC & 99.85 $\pm$ 0.0040 &
\textcolor{darkgreen}{\textbf{+0.09\%}} & $1.30 \times 10^{-3}$ & 0.6797 \\
\hline
Corel5k &
SWIN-TF & 98.01 $\pm$ 0.0323 &
VIT-B16 & SWIN-TF & RDPAC & 98.52 $\pm$ 0.0103 &
\textcolor{darkgreen}{\textbf{+0.52\%}} & $6.00 \times 10^{-8}$ & 1.2836 \\
\hline
CUB200 &
SWIN-TF & 82.05 $\pm$ 0.0067 &
VIT-B16 & SWIN-TF & RDPAC & 83.33 $\pm$ 0.0376 &
\textcolor{darkgreen}{\textbf{+1.56\%}} & $7.55 \times 10^{-10}$ & 4.5326 \\
\hline
\end{tabular}%
}
\end{table*}

\end{document}